\documentclass[11pt]{article}

\usepackage[margin=1in]{geometry}

\usepackage{enumitem}
\usepackage{multirow}
\usepackage{braket}
\usepackage{float}
\usepackage{listings}
\usepackage{xcolor}
\usepackage{booktabs}
\usepackage{tabularx} 
\usepackage{ragged2e}
\newcolumntype{Y}{>{\RaggedRight\arraybackslash}X}
\usepackage{amsmath, amsthm, mathtools, amssymb}
\usepackage[numbers]{natbib}  
\usepackage{graphicx}
\usepackage{hyperref}
\usepackage{url}
\usepackage{authblk}  

\hypersetup{
    colorlinks=true,
    linkcolor=blue,
    citecolor=blue,
    urlcolor=blue
}

\begin{document}

\title{An Invariant Compiler for Neural ODEs in \\ AI-Accelerated Scientific Simulation}

\author[1]{Fangzhou Yu}
\author[1]{Yiqi Su}
\author[1]{Ray Lee}
\author[1]{Shenfeng Cheng}
\author[1]{Naren Ramakrishnan}
\affil[1]{Virginia Tech, VA, USA}
\affil[ ]{\texttt{\{yfangz7, yiqisu, ray.lee, chengsf, naren\}@vt.edu}}

\date{}  

\maketitle

\begin{abstract}
Neural ODEs are increasingly used as continuous-time models for scientific and sensor data, but unconstrained neural ODEs can drift and violate domain invariants (e.g., conservation laws), yielding physically implausible solutions. In turn, this can compound error in long-horizon prediction and surrogate simulation. Existing solutions typically aim to enforce invariance by soft penalties or other forms of regularization, which can reduce overall error but do not guarantee that trajectories will not leave the constraint manifold. We introduce the invariant compiler, a framework that enforces invariants by construction: it treats invariants as first-class types and uses an LLM-driven compilation workflow to translate a generic neural ODE specification into a structure-preserving architecture whose trajectories remain on the admissible manifold in continuous time (and up to numerical integration error in practice). This compiler view cleanly separates what must be preserved (scientific structure) from what is learned from data (dynamics within that structure). It provides a systematic design pattern for invariant-respecting neural surrogates across scientific domains.
\end{abstract}


\section{Introduction}

Learning dynamical systems from data is a cornerstone of modern science and engineering, with applications ranging from discovering physical laws \cite{Brunton2016SINDy} to building digital twins for complex industrial processes \cite{Kapteyn2021DigitalTwins}. Neural Ordinary Differential Equations (NODEs) have emerged as a powerful, general-purpose framework for this task, parameterizing the vector field of a system's dynamics with a neural network \cite{Chen2018NeuralODE}. However, a fundamental challenge remains: standard data-driven models are often ``physics-agnostic.'' When deployed for long-horizon prediction, their trajectories can drift, violating fundamental physical principles such as conservation of energy, mass, or momentum, leading to physically nonsensical and unreliable predictions \cite{Greydanus2019HNN, Lutter2019DeepLagrangian}.

Significant research has focused on embedding physical priors into neural network models to mitigate these issues. These efforts can be broadly categorized into two paradigms. The first approach enforces invariants through \textbf{soft penalties} in the loss function, encouraging the model to respect a known conserved quantity \cite{Raissi2019PINN,Sholokhov2023PINODE,Hao2024SymmetryRegularizedNeuralODE}. While simple to implement, this provides no formal guarantees; the learned dynamics can and do drift from the constraint manifold over time.

\begin{figure}[H]
    \centering
    \includegraphics[width=0.95\textwidth]{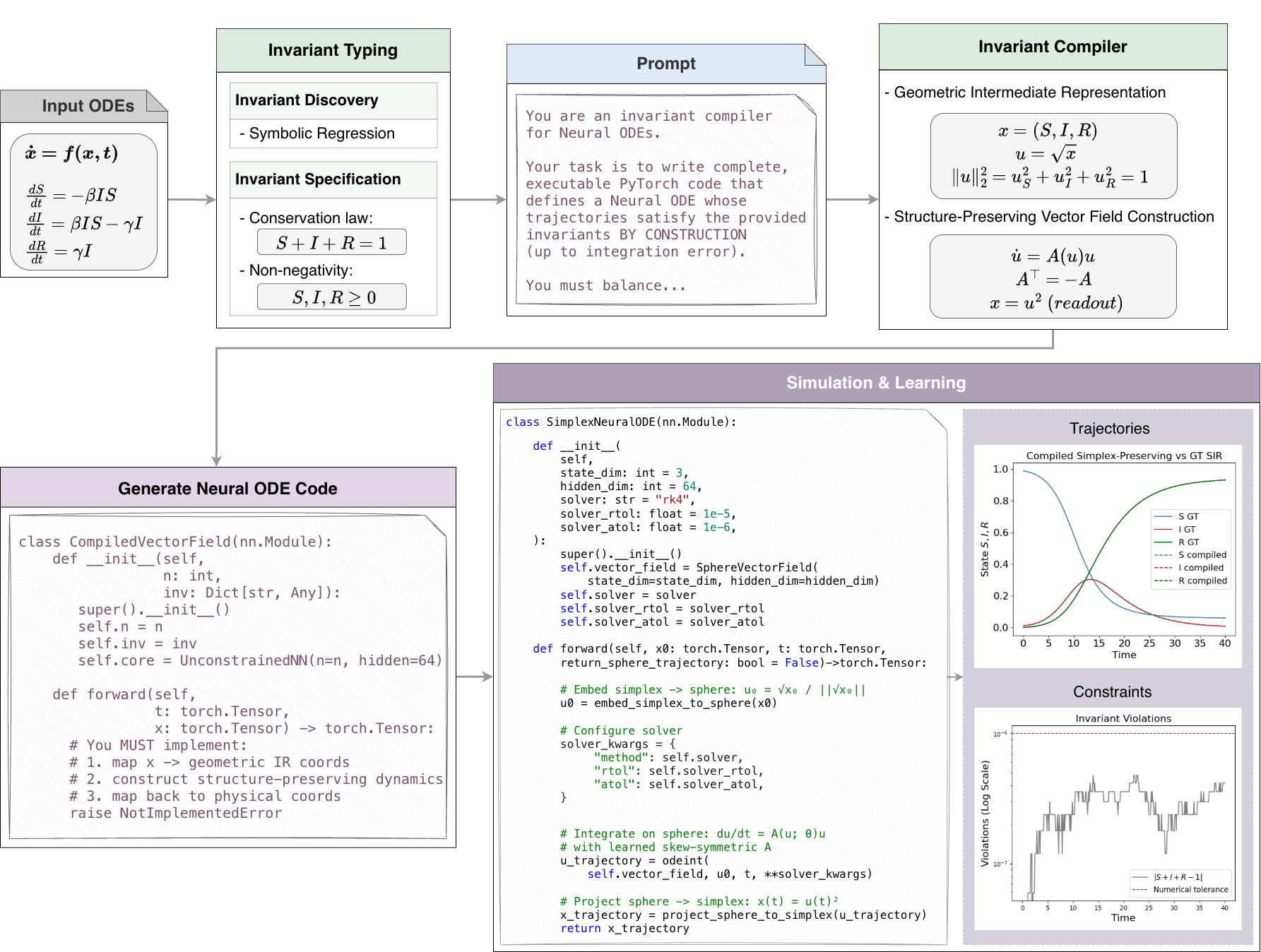}
    \caption{Invariant Compiler Pipeline for Neural ODEs. Epidemiological SIR dynamics are used as a representative example.}
    \label{fig:compiler}
\end{figure}

The second paradigm seeks to enforce constraints \textbf{exactly} through architectural or algorithmic \cite{dupont2019augmented,finzi2020simplifying}. This includes projection-based methods that correct states onto valid manifolds \cite{PNDE2024, Pal2025SemiExplicitNeuralDAE, Lou2020NeuralManifoldODE, Falorsi2020NeuralODEManifolds}, energy-preserving architectures derived from Hamiltonian or Lagrangian mechanics \cite{Greydanus2019HNN, Cranmer2020LNN, Lutter2019DeepLagrangian, Desai2021PHNN, cheng2024learning, Zhang2022GFINNs}, and networks enforcing stoichiometric or symmetry constraints \cite{Ji2021CRNN, Doppell2024AtomConservingCRNN, Kircher2024GRNN, Finzi2021EquivariantMLP, Satorras2021EnEquivariantFlows}. While effective for their targeted invariants, these methods exist as a fragmented collection of point solutions, each requiring bespoke design for each new type of constraint. There lacks a unifying perspective that treats physical invariants not as special cases, but as a fundamental aspect of the model's type signature. This fragmentation requires researchers to invent a new architecture for each new type of constraint, hindering the rapid application of structured models to new scientific domains.

In this work, we propose an {\bf Invariant Compiler} framework to bridge this gap via an LLM-driven compilation workflow. We argue that invariants should be treated as first-class types that can be systematically compiled into the structure of a Neural ODE. Our framework separates \emph{what} must be preserved (the structure, specified by the invariant's type) from \emph{what} must be learned (the dynamics within that structure). The compiler takes a generic NODE vector field and systematically rewrites it into an architecture whose trajectories are guaranteed to remain on the admissible manifold in continuous time. We demonstrate the unifying power of this view by showing how a broad class of existing and novel structure-preserving architectures, e.g., from Hamiltonian mechanics to stoichiometric constraints, arise as instances of this compiler pipeline.

Our key contributions are: (i) the invariant compiler design, (ii) a cataloging of invariants across multiple domains of physics, science, and engineering, and (iii) empirical results demonstrating constraint satisfaction, forecasting and extrapolation accuracy, and fidelity to underlying physical structure and expected qualitative dynamics.

\section{Invariant Compiler Design}

We treat invariants as first-class types.
A key design principle of the compiler is to separate what must be preserved from what must be learned.
Once an invariant type is specified, either from domain knowledge or discovered from data, the compiler deterministically produces a structure-preserving vector field whose trajectories remain on the valid state manifold for all time. As shown in Fig.~\ref{fig:compiler}, the compilation workflow consists of the following stages.

\paragraph{Stage 1: Input ODEs.}
The pipeline begins with an input dynamical system specification, e.g., $\dot{x}=f(x,t)$, provided in symbolic form or as executable code. This reference ODE defines the intended dynamics but is not required to satisfy invariants exactly~\cite{marsden2001discrete,hairer2006geometric}. It serves as the semantic target that the compiled model aims to preserve while enforcing additional structure.

\paragraph{Stage 2: Invariant Typing (Discovery or Specification).}
Invariants are obtained through two complementary mechanisms.
(i) \emph{Invariant discovery}: when invariants are unknown, data-driven methods such as symbolic regression are used to propose candidate conserved quantities or constraints~\cite{sciadv.aay2631, pnas.1906995116, xiang2025physicsconstrained, SINDy-BVP}.
(ii) \emph{Invariant specification}: when invariants are known, users explicitly provide constraints such as conservation laws, non-negativity, ordering constraints, energetic structure, or stoichiometric balances~\cite{Greydanus2019HNN, Desai2021PHNN, Oettinger2018GENERICIntegrators, M_ller_2012,feinberg2019foundations}.
Together, these invariants define the admissible state manifold on which the dynamics must evolve.

\paragraph{Stage 3: Prompt (Program Specification).}
The input ODEs and the typed invariants are assembled into a structured prompt that serves as the compiler's program specification. The prompt encodes the base dynamics, the invariants to be preserved, and hard compilation constraints (e.g., invariants must hold by construction, without penalties or post-hoc projection). This step makes the interface between user intent and the compiler explicit, while the downstream compilation remains deterministic and type-driven~\cite{epiagent}. Table~\ref{tab:prompt-taxonomy} summarizes the prompt variants supported by the invariant compiler and their corresponding invariant types and structure-preserving rewrite rules.

\paragraph{Stage 4: Invariant Compiler.}
The invariant compiler interprets the prompt and performs two tightly coupled steps:
(i) \emph{Geometric Intermediate Representation (IR)}: each invariant type is mapped to a geometric or algebraic representation in which the constraint becomes structural (e.g., simplices to spheres, conservation laws to null spaces, PSD constraints to Cholesky manifolds, energy conservation to symplectic or Poisson manifolds).
(ii) \emph{Structure-Preserving Vector Field Construction}: the vector field is rewritten into a form that is tangent to the admissible manifold, using invariant-specific constructions such as skew-symmetric generators, null-space parameterizations, factorized dynamics, Hamiltonian/Poisson formulations, or tangent-cone projections.
This stage constitutes the core compilation step and guarantees exact invariance in continuous time.

\paragraph{Stage 5: Generate Neural ODE Code.}
The compiler emits executable Neural ODE code (e.g., a PyTorch \texttt{nn.Module} implementing \texttt{forward(t,x)}), in which the invariant-preserving structure is fixed and only unconstrained submodules (e.g., neural networks producing interaction terms, matrices, or reaction rates) are learnable. The generated code also includes invariant diagnostics for verification.

\paragraph{Stage 6: Simulation \& Learning.}
Finally, the generated Neural ODE is executed in a standard run-time loop using an ODE solver and, when applicable, trained on data. Because invariants are enforced by construction in the compiled vector field, simulation and learning proceed without penalties, normalization, or post-hoc projection, up to numerical integration error.

\section{Catalog of Invariant Types}
\label{sec:catalog}

Table~\ref{tab:catalog} summarizes the invariant types supported by the compiler, along with their geometric representations, enforcement mechanisms, and representative applications. We describe each type below; full mathematical derivations and formal verification proofs appear in Appendices~\ref{app:simplex}--\ref{app:first_integral}.

\subsection{Simplex Preservation via Skew-Symmetric Dynamics}
\label{sec:simplex}

Systems whose states represent proportions or frequencies---such as compartmental epidemiological models (SIR, SEIR) \cite{kermack1927contribution,DBLP:conf/kdd/LiWC24, DBLP:conf/kdd/Dang0XBSCGSCM23}, market shares \cite{kimball1957industrial}, or allele frequencies \cite{wright1931mendelian}---must remain on the probability simplex \cite{sandholm2015population} $\Delta^{n-1} = \{x \in \mathbb{R}^n : \sum_i x_i = 1,\; x_i \geq 0\}$.

We enforce simplex invariance via a three-step construction inspired by the analogy between probability simplices and quantum state spaces \cite{beneduci2021unifying,leon2007quantum}. First, we embed the simplex into the unit sphere $S^{n-1}$ via the square-root map $u_i = \sqrt{x_i}$, so that $\|u\|_2^2 = \sum_i x_i = 1$. Second, we evolve $u(t)$ on the sphere using skew-symmetric dynamics:
\begin{equation}
\frac{du}{dt} = A(u)\, u, \quad A(u) = \tfrac{1}{2}(f_\theta(u) - f_\theta(u)^\top),
\end{equation}
where $f_\theta$ is a neural network. Since $A^\top = -A$, we have $u^\top A u = 0$, which preserves $\|u\|_2 = 1$. Third, we recover simplex coordinates via squaring: $x_i(t) = u_i(t)^2$. Non-negativity is automatic ($u_i^2 \geq 0$), and the sum constraint follows from norm preservation ($\sum_i u_i^2 = 1$). See Appendix~\ref{app:simplex} for the full derivation.

\subsection{Cone Invariants via Tangent-Space Projection}
\label{sec:cones}

Many physical systems evolve within conic regions of state space---sets closed under positive scaling. Two cones arise frequently.

\paragraph{Lorentz Cone.} Given a system
\begin{equation}
    \frac{dz}{dt} = f(z),
\end{equation}
the constraint $z \in \mathcal{L}^{n+1}$ appears in relativistic causal structure \cite{TaylorWheeler1992SpacetimePhysics} and robust optimization \cite{Boyd2004Convex}, where the Lorentz cone $\mathcal{L}^{n+1}$ is defined by
\begin{equation}
    \mathcal{L}^{n+1} = \left\{ (t, x) \in \mathbb{R} \times \mathbb{R}^n : t \geq \|x\|_2 \right\}.
\end{equation}
Forward invariance is enforced by projecting the learned neural vector field $\hat{f}_\theta$ onto the tangent cone at each point \cite{white2024projectedneuraldifferentialequations}:
\begin{equation}
\frac{dz}{dt} = f_\theta(z) \coloneqq \pi_{T_K(z)}(\hat{f}_\theta(z)), \quad z(0) \in K,
\end{equation}
where $K \coloneqq \mathcal{L}^{n+1}$ and $\pi_{T_K(z)}$ denotes orthogonal projection onto the tangent cone $T_K(z)$. The viability condition $f_\theta(z) \in T_K(z)$ guarantees that trajectories starting in $K$ remain in $K$ \cite{Aubin1991Viability}. The projection has closed-form expressions for interior, boundary, and apex cases (Appendix~\ref{app:lorentz}).

\paragraph{PSD Cone.} The constraint $P \succeq 0$ arises in covariance dynamics and filtering \cite{Anderson2005OptimalFiltering}. Rather than projecting onto $\mathcal{S}_+^n$ (requiring eigenvalue decomposition), we employ a Cholesky parameterization \cite{Bierman1977Factorization,Ko2025CholeskyKalmanNet}: evolve a lower-triangular factor $L(t)$ via $\frac{dL}{dt} = g_\theta(L)$ and recover $P(t) = L(t)L(t)^\top$. Symmetry and positive semidefiniteness hold automatically since $v^\top LL^\top v = \|L^\top v\|^2 \geq 0$ (Appendix~\ref{app:psd}).

\begin{table}
\caption{Catalog of invariant types supported by the Invariant Compiler. Each row shows one invariant type with its constraint, the geometric or algebraic mechanism used to enforce it, and representative application domains. Full derivations and proofs appear in the referenced appendix.}
\label{tab:catalog}
\centering
\footnotesize
\setlength{\tabcolsep}{3pt}
\begin{tabular}{@{}p{1.8cm}p{2.4cm}p{2.0cm}p{3.2cm}p{2.4cm}p{1.8cm}c@{}}
\toprule
\textbf{Invariant Type} & \textbf{Constraint} & \textbf{Geometric IR} & \textbf{Enforcement Mechanism} & \textbf{Applications} & \textbf{Key Refs.} & \textbf{App.} \\
\midrule
Simplex / Norm
  & $\sum_i x_i = 1$, $x_i \geq 0$
  & Unit sphere $S^{n-1}$ via $\sqrt{\cdot}$ map
  & Skew-sym.\ dynamics: $\dot{u} = A(u)u$, $A^\top = -A$; $x_i = u_i^2$
  & Epidemiology, market share, pop.\ genetics
  & \cite{beneduci2021unifying,leon2007quantum}
  & \ref{app:simplex} \\
\addlinespace
Lorentz Cone
  & $t \geq \|x\|_2$
  & Lorentz cone $\mathcal{L}^{n+1}$
  & Tangent-cone proj.: $\dot{z} = \pi_{T_K(z)}(\hat f_\theta(z))$
  & Relativity, robust opt., signal proc.
  & \cite{Aubin1991Viability,TaylorWheeler1992SpacetimePhysics, Hauswirth2020ObliqueProjectedDynamicalSystems}  
  & \ref{app:lorentz} \\
\addlinespace
PSD Cone
  & $P \succeq 0$
  & Cholesky manifold
  & Cholesky param.: $P = LL^\top$, evolve $L$
  & Covariance dynamics, Kalman filtering
  & \cite{Bierman1977Factorization,Ko2025CholeskyKalmanNet}
  & \ref{app:psd} \\
\addlinespace
Center of Mass
  & $\sum_i m_i \mathbf{r}_i = \mathbf{0}$, $\sum_i m_i \mathbf{v}_i = \mathbf{0}$
  & Null space of mass-weighted sum
  & Mean subtraction: $\dot{\mathbf{r}}_i = \mathbf{v}_i - \bar{\mathbf{v}}$, $\dot{\mathbf{v}}_i = \mathbf{a}_i - \bar{\mathbf{a}}$
  & $N$-body, molecular dynamics
  & \cite{PrantlUKT22,BattagliaPLRK16}
  & \ref{app:com} \\
\addlinespace
Stoichiometric
  & $\mathbf{M}  c(t) = \text{const}$
  & Null space of molecular matrix $\mathbf{M}$
  & Null-space proj.: $\dot{c} = B \cdot r_\theta(c,t)$, $\mathbf{M}B = 0$
  & Chemical kinetics, metabolic networks
  & \cite{Kircher2024GRNN,Doppell2024AtomConservingCRNN}
  & \ref{app:stoich} \\
\addlinespace
Hamiltonian
  & $\frac{dH}{dt} = 0$, Jacobi identity
  & Symplectic / Poisson manifold
  & Latent canonical coords via INN: $\dot{z} = J_0 \nabla_z K(z)$
  & Pendulum, Lotka-Volterra, rigid body
  & \cite{jin2023learning}
  & \ref{app:poisson} \\
\addlinespace
Port-Hamil.
  & $\frac{dH}{dt} \leq 0$
  & Poisson $+$ PSD dissipation
  & $\dot{z} = [J_0 - R(z)] \nabla_z K(z)$, $R = LL^\top \succeq 0$
  & Damped oscillators, circuits
  & \cite{cheng2024learning}
  & \ref{app:port_ham} \\
\addlinespace
GENERIC
  & $\frac{dH}{dt} = 0$, $\frac{dS}{dt} \geq 0$
  & Poisson $+$ proj.\ dissipation
  & Casimir-dep.\ entropy $+$ proj.\ $M_z$
  & Thermomech., polymer relaxation
  & \cite{Zhang2022GFINNs,Sipka2021LearningGENERIC}
  & \ref{app:generic} \\
\addlinespace
First Integral
  & $V(u(t)) = \text{const}$ (learned)
  & Learned constraint manifold
  & Tangent-space proj.: $(I - \nabla V^\top(\nabla V \nabla V^\top)^+ \nabla V)\hat{f}(u)$
  & Unknown conservation laws
  & \cite{FINDE2023,PNDE2024}
  & \ref{app:first_integral} \\
\bottomrule
\end{tabular}
\end{table}

\subsection{Center of Mass Frame Constraints}
\label{sec:com}

For isolated $n$-body systems, conservation of total momentum implies invariance of the center of mass frame: $\sum_i m_i \mathbf{r}_i = \mathbf{0}$ and $\sum_i m_i \mathbf{v}_i = \mathbf{0}$ \cite{PrantlUKT22,BattagliaPLRK16,BishnoiJRK24}.  These constraints need not be hard compiled into the neural ODE architecture; we include them as a useful inductive bias
that is often encountered in real systems.
Given predicted accelerations $\mathbf{a}_i = f_\theta(\mathbf{r}, \mathbf{v})_i$, we enforce these constraints by subtracting the mass-weighted mean:
\begin{equation}
\frac{d\mathbf{r}_i}{dt} = \mathbf{v}_i - \bar{\mathbf{v}}, \qquad \frac{d\mathbf{v}_i}{dt} = \mathbf{a}_i - \bar{\mathbf{a}},
\end{equation}
where $\bar{\mathbf{v}} = \frac{\sum_k m_k \mathbf{v}_k}{\sum_k m_k}$ and $\bar{\mathbf{a}} = \frac{\sum_k m_k \mathbf{a}_k}{\sum_k m_k}$. This projects out any net translation or force, ensuring the neural network's output affects only relative (internal) dynamics (Appendix~\ref{app:com}).

\subsection{Stoichiometric Constraints via Null-Space Projection}
\label{sec:stoich}

Chemical and biological systems obey elemental conservation: atoms are rearranged but never created or destroyed \cite{Feinberg2019}. For species concentrations $c(t)$, these laws take the form $\mathbf{M}  c(t) = \mathbf{M} c(0)$, where the molecular matrix $\mathbf{M}$ encodes element-to-species counts. We parameterize dynamics via the null-space basis $B$ of $\mathbf{M}$ \cite{Kircher2024GRNN,Doppell2024AtomConservingCRNN}:
\begin{equation}
\frac{dc}{dt} = B  r_\theta(c, t),
\end{equation}
where $r_\theta : \mathbb{R}^n \times \mathbb{R} \to \mathbb{R}^k$ is a neural network and $k = \dim(\mathrm{Null}(\mathbf{M}))$. Since $\mathbf{M}B = 0$ by construction, we have $\mathbf{M} \dot{c} = 0$, guaranteeing exact elemental conservation regardless of the network output (Appendix~\ref{app:stoich}).

\subsection{Hamiltonian / Poisson Structure}
\label{sec:poisson}

Many physical systems conserve energy exactly when dissipation is absent \cite{Goldstein2002ClassicalMechanics}. The dynamics take the form $\dot{u} = J(u) \nabla_u H(u)$, where $H$ is the Hamiltonian \cite{zhong2024symplecticodenetlearninghamiltonian, Greydanus2019HNN} and the skew-symmetric matrix $J(u)$ defines a Poisson structure satisfying the Jacobi identity \cite{Arnold1989MathematicalMethods}.

Rather than learning a state-dependent $J(u)$ subject to the nonlinear Jacobi identity constraint, following \citet{jin2023learning}, we adopt a latent-space approach\cite{rubanova2019latent}. An invertible neural network $g_{\theta} : u \leftrightarrow z = (q, p, c)$ maps physical coordinates to latent canonical coordinates, where the dynamics become:
\begin{equation}
\frac{dz}{dt} = J_0 \nabla_z K(z), \quad J_0 = \begin{pmatrix} 0 & I_d & 0 \\ -I_d & 0 & 0 \\ 0 & 0 & 0 \end{pmatrix},
\end{equation}
with $K(z) := H(g_\theta^{-1}(z))$. Energy conservation follows from skew-symmetry ($(\nabla K)^\top J_0 (\nabla K) = 0$), and the Jacobi identity holds automatically for the constant canonical matrix $J_0$. Casimir coordinates $c$ are frozen by the zero block, encoding additional conserved quantities. The physical Poisson matrix $J(u)$ is induced via the coordinate transformation law, inheriting the Jacobi identity from $J_0$ (Appendix~\ref{app:poisson}).

\subsection{Port-Hamiltonian (Dissipative) Structure}
\label{sec:port_ham}

Systems with friction, damping, or irreversible losses satisfy $\frac{dH}{dt} \leq 0$ \cite{Duindam2009PortHamiltonianBook}. Following \citet{cheng2024learning}, we augment the canonical latent-space dynamics with a positive semi-definite dissipation matrix $R_z$:
\begin{equation}
\frac{dz}{dt} = [J_0 - R_z(z)] \nabla_z K(z),
\end{equation}
where $R_z(z) = L_\theta(z)L_\theta(z)^\top \succeq 0$ and the mapping $L_\theta(z)$ is parameterized by a neural network. The energy rate becomes $\frac{dK}{dt} = -(\nabla_z K(z))^\top R_z(z) (\nabla_z K(z)) \leq 0$, guaranteeing passive stability by construction (Appendix~\ref{app:port_ham}).

\subsection{GENERIC / Thermodynamic Structure}
\label{sec:generic}

Thermodynamic systems exhibit both reversible (energy-conserving) and irreversible (entropy-producing) dynamics, governed by the GENERIC formalism \cite{Ottinger2005BeyondEquilibriumThermodynamics}: $\dot{u} = J \nabla H + M \nabla S$, subject to degeneracy conditions $J \nabla S = 0$ and $M \nabla H = 0$. Here, $S$ denotes the entropy functional and $M \succeq 0$ is a positive semi-definite friction matrix governing irreversible dissipation. We enforce these in latent space via two architectural choices \cite{Zhang2022GFINNs, Sipka2021LearningGENERIC}:
\begin{enumerate}[leftmargin=*,nosep]
\item \textbf{Casimir-dependent entropy}: $S(z) = \tilde{S}(c)$, so $\nabla_z S = (0, 0, \nabla_c \tilde{S})^\top$ lies in the null space of $J_0$.
\item \textbf{Projected dissipation}: $M_z = P_K \widehat{M}_z P_K$, where $P_K = I - \frac{\nabla K (\nabla K)^\top}{\|\nabla K\|^2 + \varepsilon}$ projects onto constant-energy surfaces.
\end{enumerate}
Together, these guarantee energy conservation ($\frac{dH}{dt} = 0$), non-negative entropy production ($\frac{dS}{dt} \geq 0$), and both degeneracy conditions---all by construction (Appendix~\ref{app:generic}).

\subsection{First Integral Preservation via Manifold Projection}
\label{sec:first_integral}

When conserved quantities $V(u) \in \mathbb{R}^m$ are unknown, following \citet{FINDE2023}, we learn them as neural networks $V_i(u; \phi_i)$ ($1 \leq i \leq m)$ and, following \citet{PNDE2024}, project an unconstrained base vector field $\hat{f}(u)$ onto the tangent space of the learned constraint manifold \cite{Hauswirth2020ObliqueProjectedDynamicalSystems}:
\begin{equation}
\frac{du}{dt} = (I - P(u))\,\hat{f}(u), \quad P = \nabla V^\top(\nabla V \nabla V^\top)^+ \nabla V,
\end{equation}
where $(\cdot)^+$ is the Moore--Penrose pseudo-inverse. The pseudo-inverse gracefully handles redundant constraints: if the network discovers linearly dependent conserved quantities, the projection remains well-defined. Since $(I-P)$ projects onto $\mathrm{null}(\nabla V)$, we have $\nabla V_i^\top (I-P) = 0$, ensuring all first integrals are exactly preserved (Appendix~\ref{app:first_integral}).

\section{Experiments}
\label{sec:experiments}

\subsection{Setup}
\label{sec:experimental_setup}

We evaluate the Invariant Compiler across dynamical systems spanning geometric, algebraic, energetic, and thermodynamic invariants (Table~\ref{tab:ode_catalog}); full equations and parameters are in Appendix~\ref{app:ode_systems}.

\begin{table*}[htbp]
\centering
\caption{Catalog of dynamical systems used in the experiments. Each system is designated with its dimensionality, the invariant type, and the compiled architecture used to enforce the invariant.}
\label{tab:ode_catalog}
\resizebox{\textwidth}{!}{%
\begin{tabular}{@{}lclll@{}}
\toprule
\textbf{System} & \textbf{Dim.} & \textbf{Invariant type} & \textbf{Architecture} & \textbf{Key challenge} \\
\midrule
SIR epidemiological model & 3 & Simplex ($S{+}I{+}R{=}1$) & Skew-symmetric & Population conservation \\
NOx reaction network & 5 & Stoichiometric ($\mathbf{S}\dot{\mathbf{c}}{=}\mathbf{0}$) & Null-space & Nonlinear kinetics, product inhibition \\
Chemical reaction network & 6 & Stoichiometric ($\mathbf{S}\dot{\mathbf{c}}{=}\mathbf{0}$) & Null-space & Element conservation (C, H, O) \\
Lorentz cone spiral & 3 & Cone ($t \geq \|\mathbf{x}\|_2$) & Projection & Boundary dynamics \\
Coupled radial-angular dynamics & 3 & Cone ($t \geq \|\mathbf{x}\|_2$) & Projection & Nonlinear radial-angular coupling \\
Replicator-mutator ($N{=}5$) & 5 & Simplex ($\sum x_i{=}1$) & Skew-symmetric & Diverse dynamical regimes \\
Lotka-Volterra predator-prey & 2 & Hamiltonian (Poisson) & Poisson INN & Non-canonical bracket \\
Damped harmonic oscillator & 2 & Port-Hamiltonian ($\dot{E} \leq 0$) & PH-INN & Energy dissipation \\
Thermomechanical system & 3 & GENERIC ($\dot{E}{=}0$, $\dot{S}{\geq}0$) & GENERIC INN & Coupled conservation + production \\
Extended pendulum & 3 & Poisson (Casimir + Hamiltonian) & Poisson INN & Multiple simultaneous invariants \\
Two-body gravitational & 8 & Mixed (known + learned) & FINDE & Heterogeneous invariant composition \\
\bottomrule
\end{tabular}%
}
\end{table*}

\subsubsection{Research Questions}

We evaluate the Invariant Compiler framework through experiments designed to address the following research questions, each tested on a primary system (bold) and, where applicable, a supplementary system whose results appear in Appendix~\ref{app:supplementary}:

\begin{itemize}[leftmargin=*]
    \item[\textbf{Q1}] \textbf{Constraint satisfaction.} Do compiled architectures enforce algebraic constraints to numerical precision? Primary: \textbf{NOx reaction network}. Supplementary: SIR model, chemical reaction network.
    \item[\textbf{Q2}] \textbf{Prediction accuracy.} Does encoding constraints improve or degrade prediction accuracy? Primary: \textbf{coupled radial-angular dynamics}. Supplementary: Lorentz cone spiral.
    \item[\textbf{Q3}] \textbf{Long-horizon extrapolation.} Do compiled architectures maintain bounded errors at $2\times$--$10\times$ the training horizon? Primary: \textbf{thermomechanical system}. Supplementary: damped harmonic oscillator.
    \item[\textbf{Q4}] \textbf{Conservation law fidelity.} Do learned invariants track theoretically predicted quantities? Primary: \textbf{Lotka-Volterra system}.
    \item[\textbf{Q5}] \textbf{Qualitative behaviour.} Do compiled architectures reproduce the correct dynamical regime? Primary: \textbf{Replicator-mutator system}.
    \item[\textbf{Q6}] \textbf{Scalability.} Do compiled constraints remain effective with multiple simultaneous invariants and higher structural complexity? Primary: \textbf{extended pendulum}.
    \item[\textbf{Q7}] \textbf{Composability.} Can heterogeneous invariant types---some known analytically, others learned---be composed within a single architecture? Primary: \textbf{two-body gravitational system}.
\end{itemize}

\subsubsection{Training, Architecture, Baselines, and Metrics}

\paragraph{Integration and Training.}
All models employ fourth-order Runge-Kutta (RK4) integration during both training and inference. We use multi-step rollouts with exponentially decaying weights:
\begin{equation}
    \mathcal{L} = \frac{1}{n_{\text{steps}}} \sum_{k=1}^{n_{\text{steps}}} \frac{1}{k} \left\| \hat{u}_{t+k} - u_{t+k} \right\|^2,
\end{equation}
with $n_{\text{steps}} = 4$, emphasising accurate short-term prediction while penalising long-horizon drift. Models are optimised using AdamW with learning rate $10^{-3}$, weight decay $10^{-5}$, and cosine annealing over 300--1000 epochs depending on the task. Gradient norms are clipped to 1.0 for stability.

\paragraph{Network Architecture.}
All methods share identical network capacity where applicable. For direct constraint enforcement experiments, we employ 3-layer MLPs with 64 hidden units and SiLU (or Softplus for chemical systems) activations. For structure-learning experiments, we additionally use:
\textbf{Invertible Neural Network (INN):} 8 coupling blocks using the AllInOneBlock architecture from FrEIA~\cite{freia}, with soft permutations between blocks. Each coupling block uses a 3-layer MLP subnet with hidden dimension 64 and SiLU activations.
\textbf{Dissipation networks:} 3-layer MLPs outputting Cholesky factors, initialised with small weights ($\sigma = 0.01$) for stability.

\paragraph{LLM Implementation.}
The invariant compiler uses a large language model (LLM) as a program synthesis and rewrite engine, rather than as a predictive model. Given the prompt specification (Listing~\ref{lst:prompt}), the LLM generates executable PyTorch code that implements the required geometric intermediate representation and structure-preserving vector field. All compilation logic is driven by the prompt and deterministic post-processing; the LLM is not trained or fine-tuned on the target dynamical systems. Unless otherwise stated, we use a standard GPT-class model with temperature set to zero to ensure deterministic code generation. 

\paragraph{Baselines.}
All experiments compare against an \textbf{Unconstrained (Neural ODE)} baseline: a vanilla Neural ODE that directly parameterises the vector field $\dot{x} = f_\theta(x)$ without any structural constraints. The two experimental sections employ different soft-constraint baselines suited to their respective settings.
For constraint satisfaction experiments (Sections~\ref{sec:q1}--\ref{sec:q2}), we additionally compare against a \textbf{Penalty} baseline trained with an additional penalty term $\lambda \cdot \mathcal{L}_{\text{constraint}}$ ($\lambda = 10$).
For structure-learning experiments (Sections~\ref{sec:q3}--\ref{sec:q4}), we instead compare against a \textbf{PINNs-style} baseline that augments the Neural ODE loss with physics-informed penalty terms derived from the \emph{known} analytical form of the system's invariants.

\paragraph{Metrics.}
We evaluate models using trajectory MSE (reported separately for training window, extrapolation window, and total) and constraint-specific measures. For systems with explicit constraints, we report mean violation; for systems with conserved quantities, we measure deviation from theoretical values:
\begin{align}
    \text{Deviation} &= \frac{1}{T} \int_0^T |Q(t) - Q_{\text{theory}}(t)| \, dt.
\end{align}
To make relative performance transparent, we additionally report \emph{improvement factors} (IF): the ratio of the best baseline metric to the compiled-architecture metric.

\subsection{Q1: Constraint Satisfaction}
\label{sec:q1}

We evaluate whether compiled architectures can enforce hard algebraic constraints to numerical precision, using the NOx reaction network---a system with strongly nonlinear kinetics including product inhibition, substrate inhibition, and fractional-power terms. We train on 1000 trajectories ($T_{\text{end}}{=}10$, 200 points per trajectory) and evaluate on 200 held-out initial conditions. Results for the supplementary systems (SIR model, six-species chemical network) appear in Appendix~\ref{app:supplementary}, Tables~\ref{tab:sir_results}--\ref{tab:chem_results}.

The stoichiometric matrix $\mathbf{S} \in \mathbb{R}^{2 \times 5}$ encodes element-to-species relationships for nitrogen and oxygen. Following Section~\ref{sec:stoich}, we parameterise dynamics via the null-space basis, ensuring $\mathbf{S} \cdot \dot{\mathbf{c}} = \mathbf{0}$ by construction.

\begin{table}[htbp]
\centering
\caption{NOx reaction network (\textbf{Q1}: constraint satisfaction). Mean $\pm$ std over 200 test trajectories. IF = improvement factor of our method over the best baseline.}
\label{tab:nox_results}
\footnotesize
\setlength{\tabcolsep}{4pt}
\begin{tabular}{@{}lcccc@{}}
\toprule
\textbf{Metric} & \textbf{Unconst.} & \textbf{Penalty} & \textbf{Ours} & \textbf{IF} \\
\midrule
MSE (Train) & $5.86\text{e-}7${\scriptsize $\pm 6.10\text{e-}7$} & $1.02\text{e-}2${\scriptsize $\pm 1.02\text{e-}2$} & $\mathbf{1.80\text{e-}7}${\scriptsize $\pm 1.48\text{e-}7$} & $3.3\times$ \\
MSE (Extrap) & $1.23\text{e-}6${\scriptsize $\pm 1.68\text{e-}6$} & $1.22\text{e-}2${\scriptsize $\pm 1.10\text{e-}2$} & $\mathbf{9.18\text{e-}8}${\scriptsize $\pm 8.39\text{e-}8$} & $13\times$ \\
MSE (Total) & $9.09\text{e-}7${\scriptsize $\pm 1.04\text{e-}6$} & $1.12\text{e-}2${\scriptsize $\pm 1.06\text{e-}2$} & $\mathbf{1.36\text{e-}7}${\scriptsize $\pm 9.19\text{e-}8$} & $6.7\times$ \\
\midrule
Elem.\ Dev. & $6.80\text{e-}3${\scriptsize $\pm 4.15\text{e-}3$} & $1.10\text{e-}4${\scriptsize $\pm 7.42\text{e-}5$} & $\mathbf{1.06\text{e-}6}${\scriptsize $\pm 8.26\text{e-}7$} & $\mathbf{104\times}$ \\
\bottomrule
\end{tabular}
\end{table}

Table~\ref{tab:nox_results} demonstrates that the null-space architecture reduces element conservation violations by \textbf{over two orders of magnitude} ($104\times$) compared to the best baseline (penalty), and by nearly \textbf{four orders of magnitude} compared to the unconstrained model. 
Element deviations remain at ${\sim}10^{-6}$ (near numerical precision), whereas soft penalties reduce violations only to ${\sim}10^{-4}$ while substantially degrading MSE.
Consistent results are observed for the simpler SIR and chemical systems (Appendix~\ref{app:supplementary}), where compiled architectures achieve $6$ and $4$ orders of magnitude violation reduction, respectively.

\begin{figure}[htbp]
  \centering
  \includegraphics[width=0.75\textwidth]{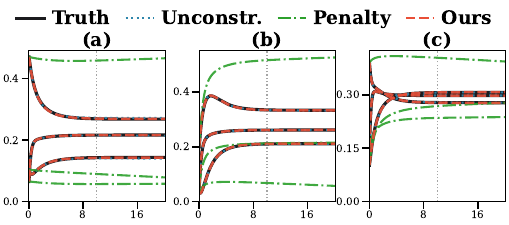}
  \caption{NOx reaction network (\textbf{Q1}). Trajectory comparison showing species concentrations over time under three different initial conditions: (a), (b), and (c). Each initial condition leads to distinct transient dynamics and converges to different equilibrium values. The vertical dashed line marks the end of the training window.}
  \label{fig:q1_nox}
\end{figure}

\subsection{Q2: Prediction Accuracy}
\label{sec:q2}

We evaluate whether compiled constraints improve or degrade prediction accuracy, using coupled radial-angular dynamics on the Lorentz cone---a system with nonlinear coupling between radial growth and angular velocity. We train on 1000 trajectories ($T_{\text{end}}{=}10$, 500 points) and evaluate on 200 held-out trajectories to $T_{\text{end}}{=}16$ ($1.6\times$ extrapolation). Results for the supplementary system (simple Lorentz cone spiral) appear in Appendix~\ref{app:supplementary}, Table~\ref{tab:lorentz_simple_results}.

The projection-based architecture guarantees that all predicted trajectories remain within the cone $\mathcal{L}^3$.

\begin{table}[htbp]
\centering
\caption{Coupled radial-angular dynamics on the Lorentz cone (\textbf{Q2}: prediction accuracy). Mean $\pm$ std over 200 test trajectories.}
\label{tab:radial_angular}
\footnotesize
\setlength{\tabcolsep}{4pt}
\begin{tabular}{@{}lcccc@{}}
\toprule
\textbf{Metric} & \textbf{Unconst.} & \textbf{Penalty} & \textbf{Ours} & \textbf{IF} \\
\midrule
MSE (Train)     & $3.04\text{e-}4${\scriptsize $\pm 2.46\text{e-}3$} & $2.45\text{e-}1${\scriptsize $\pm 8.31\text{e-}1$} & $\mathbf{1.02\text{e-}4}${\scriptsize $\pm 4.81\text{e-}4$} & $3.0\times$ \\
MSE (Extrap)    & $4.90\text{e-}4${\scriptsize $\pm 4.03\text{e-}3$} & $4.43\text{e-}1${\scriptsize $\pm 1.40\text{e}{+}0$} & $\mathbf{1.55\text{e-}4}${\scriptsize $\pm 7.55\text{e-}4$} & $3.2\times$ \\
MSE (Total)     & $3.74\text{e-}4${\scriptsize $\pm 3.05\text{e-}3$} & $3.20\text{e-}1${\scriptsize $\pm 1.05\text{e}{+}0$} & $\mathbf{1.22\text{e-}4}${\scriptsize $\pm 5.84\text{e-}4$} & $3.1\times$ \\
\midrule
Viol.\ (Mean)   & $4.13\text{e-}4${\scriptsize $\pm 4.37\text{e-}4$} & $\mathbf{1.07\text{e-}11}${\scriptsize $\pm 3.30\text{e-}11$} & $3.75\text{e-}11${\scriptsize $\pm 1.34\text{e-}10$} & --- \\
\bottomrule
\end{tabular}
\end{table}

Table~\ref{tab:radial_angular} reveals a key finding: the Lorentz PDS architecture achieves $\mathbf{3\times}$ lower trajectory MSE than the unconstrained baseline while maintaining cone violations near numerical precision (${\sim}10^{-8}$). Unlike the simple spiral (Appendix~\ref{app:supplementary}), where unconstrained models could gain slight MSE advantages by ``shortcutting'' through forbidden cone interiors, the coupled radial-angular dynamics demand accurate modelling of the interplay between radial growth and angular velocity---a task where geometric consistency provides a genuine inductive bias. The penalty baseline, while achieving low violations, catastrophically degrades prediction accuracy (MSE $> 10^{-1}$), demonstrating once more that soft-constraint regularisation fundamentally conflicts with the data-fitting objective.

\begin{figure}[htbp]
  \centering
  \includegraphics[width=0.75\textwidth]{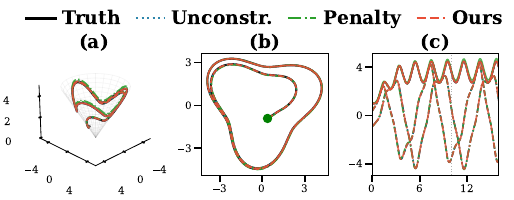}
  \caption{Coupled radial-angular dynamics on the Lorentz cone (\textbf{Q2}). 
  (a) Trajectories in 3D cone space. 
  (b) Projection onto the $(x_1, x_2)$ plane. 
  (c) Time evolution of the temporal component $t$, radial norm $|x|$, and spatial coordinates $x_1$, $x_2$, with the vertical dashed line marking the training horizon.}
  \label{fig:q2_lorentz}
\end{figure}

\subsection{Q3: Long-Horizon Extrapolation}
\label{sec:q3}

We evaluate the ability of compiled architectures to maintain accuracy over extended time horizons, using the thermomechanical system---a damped oscillator coupled to a heat bath where total energy is conserved while entropy increases monotonically. Training uses a single trajectory of 1000 points over $t \in [0, 30]$, with evaluation at $2\times$, $5\times$, and $10\times$ the training horizon. Results for the supplementary system (damped harmonic oscillator) appear in Appendix~\ref{app:supplementary}, Table~\ref{tab:damped_results}.

Table~\ref{tab:thermomech} reveals a striking pattern: at $10\times$ extrapolation, the GENERIC INN maintains MSE below $10^{-6}$ while the Neural ODE degrades to $2.1 \times 10^{-4}$---over $\mathbf{200\times}$ worse. Even the PINNs-style baseline, despite access to the true analytical invariants, degrades to $1.0 \times 10^{-5}$, more than $\mathbf{10\times}$ worse than the compiled architecture. The improvement factors grow with the extrapolation horizon, underscoring that hard structural enforcement provides compounding benefits over time: soft constraints ``leak'' gradually, whereas compiled constraints remain exact. Energy deviation at $10\times$ reaches $2.3 \times 10^{-2}$ for the Neural ODE but only $1.3 \times 10^{-3}$ for the GENERIC INN---an $\mathbf{18\times}$ improvement.

\begin{table}[htbp]
\centering
\caption{Thermomechanical system (\textbf{Q3}: long-horizon extrapolation). A single training trajectory is extrapolated to $2\times$, $5\times$, and $10\times$ the training horizon. IF at $10\times$ relative to best baseline.}
\label{tab:thermomech}
\footnotesize
\setlength{\tabcolsep}{3pt}
\begin{tabular}{@{}llcccc@{}}
\toprule
\textbf{Metric} & \textbf{Model} & \textbf{$\times 2$} & \textbf{$\times 5$} & \textbf{$\times 10$} & \textbf{IF ($\times 10$)} \\
\midrule
\multirow{3}{*}{MSE (Train)} & Unconst. & 2.91e-06 & 2.91e-06 & 2.91e-06 & \\
 & PINNs & 2.56e-06 & 2.56e-06 & 2.56e-06 & \\
 & Ours & \textbf{1.16e-06} & \textbf{1.16e-06} & \textbf{1.16e-06} & $2.2\times$ \\
\midrule
\multirow{3}{*}{MSE (Extrap)} & Unconst. & 2.41e-05 & 1.40e-04 & 2.33e-04 & \\
 & PINNs & 2.16e-06 & 6.75e-06 & 1.12e-05 & \\
 & Ours & \textbf{7.21e-07} & \textbf{8.47e-07} & \textbf{9.77e-07} & $\mathbf{11\times}$ \\
\midrule
\multirow{3}{*}{MSE (Total)} & Unconst. & 1.35e-05 & 1.13e-04 & 2.10e-04 & \\
 & PINNs & 2.36e-06 & 5.91e-06 & 1.03e-05 & \\
 & Ours & \textbf{9.40e-07} & \textbf{9.10e-07} & \textbf{9.95e-07} & $\mathbf{10\times}$ \\
\midrule
\multirow{3}{*}{Energy Dev.} & Unconst. & 4.80e-03 & 1.56e-02 & 2.29e-02 & \\
 & PINNs & \textbf{8.50e-04} & 3.26e-03 & 4.94e-03 & \\
 & Ours & 9.24e-04 & \textbf{1.11e-03} & \textbf{1.27e-03} & $\mathbf{3.9\times}$ \\
\midrule
\multirow{3}{*}{Entropy Dev.} & Unconst. & 2.18e-03 & 6.97e-03 & 1.02e-02 & \\
 & PINNs & 4.02e-04 & 1.46e-03 & 2.20e-03 & \\
 & Ours & \textbf{3.63e-04} & \textbf{4.75e-04} & \textbf{5.53e-04} & $\mathbf{4.0\times}$ \\
\bottomrule
\end{tabular}
\end{table}

\begin{figure}[htbp]
  \centering
  \includegraphics[width=0.7\textwidth]{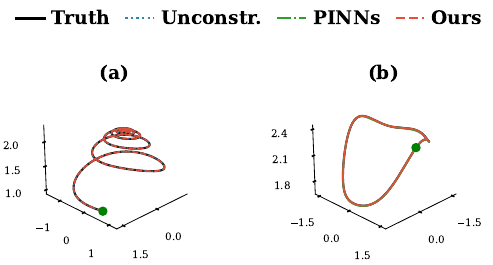}
  \caption{3D phase-space portraits. \textbf{(a)} Thermomechanical system (\textbf{Q3}) \textbf{(b)} Extended pendulum (\textbf{Q6}) Green dots indicate initial conditions.}
  \label{fig:3d_phase_space}
\end{figure}

\subsection{Q4: Conservation Law Fidelity}
\label{sec:q4}

We evaluate whether compiled architectures faithfully maintain the underlying conservation structure, using the Lotka-Volterra predator-prey system. Training data spans $t \in [0, 30]$ (single trajectory), with evaluation at $2\times$, $5\times$, and $10\times$ extrapolation.

\begin{table}[htbp]
\centering
\caption{Lotka-Volterra system (\textbf{Q4}: conservation law fidelity). IF at $10\times$ relative to best baseline.}
\label{tab:lotka_volterra}
\footnotesize
\setlength{\tabcolsep}{3pt}
\begin{tabular}{@{}llcccc@{}}
\toprule
\textbf{Metric} & \textbf{Model} & \textbf{$\times 2$} & \textbf{$\times 5$} & \textbf{$\times 10$} & \textbf{IF ($\times 10$)} \\
\midrule
\multirow{3}{*}{MSE (Train)} & Unconst. & 4.39e-06 & 4.39e-06 & 4.39e-06 & \\
 & PINNs & 7.54e-06 & 7.54e-06 & 7.54e-06 & \\
 & Ours & \textbf{7.61e-07} & \textbf{7.61e-07} & \textbf{7.61e-07} & $5.8\times$ \\
\midrule
\multirow{3}{*}{MSE (Extrap)} & Unconst. & 7.38e-05 & 1.08e-03 & 1.15e-02 & \\
 & PINNs & 6.03e-05 & 1.66e-04 & 3.47e-04 & \\
 & Ours & \textbf{2.25e-06} & \textbf{1.11e-05} & \textbf{4.76e-05} & $\mathbf{7.3\times}$ \\
\midrule
\multirow{3}{*}{MSE (Total)} & Unconst. & 3.91e-05 & 8.63e-04 & 1.04e-02 & \\
 & PINNs & 3.39e-05 & 1.34e-04 & 3.13e-04 & \\
 & Ours & \textbf{1.51e-06} & \textbf{9.00e-06} & \textbf{4.29e-05} & $\mathbf{7.3\times}$ \\
\midrule
\multirow{3}{*}{Energy Dev.} & Unconst. & 1.18e-03 & 2.14e-03 & 3.44e-03 & \\
 & PINNs & \textbf{2.14e-04} & 4.23e-04 & 6.15e-04 & \\
 & Ours & 3.80e-04 & \textbf{3.99e-04} & \textbf{4.14e-04} & $\mathbf{1.5\times}$ \\
\bottomrule
\end{tabular}
\end{table}

Table~\ref{tab:lotka_volterra} shows that the Poisson INN achieves the tightest energy deviation at both $5\times$ and $10\times$ extrapolation, despite \emph{never having access to the analytical Hamiltonian}. The learned Hamiltonian deviation remains bounded ($< 4.2 \times 10^{-4}$) even at $10\times$ extrapolation, whereas the Neural ODE's deviation grows to $3.4 \times 10^{-3}$---an $\mathbf{8.3\times}$ gap. The PINNs-style baseline, which directly penalises the known conserved quantity, achieves competitive short-horizon fidelity but falls behind the Poisson INN at $5\times$ and $10\times$, illustrating that soft penalties inevitably drift. The Poisson INN simultaneously achieves $\mathbf{7\times}$ lower trajectory MSE at $10\times$ extrapolation, confirming that structure fidelity and prediction accuracy are mutually reinforcing.

\subsection{Q5: Qualitative Behaviour}
\label{sec:q5}

We evaluate whether compiled architectures reproduce the correct qualitative dynamics---oscillatory, convergent, or transient behaviour depending on initial conditions---using the Replicator-Mutator system ($N{=}5$ species). We train on 1000 trajectories sampled from a Dirichlet distribution and evaluate on 200 held-out trajectories over $t \in [0, 20]$.

\begin{table}[htbp]
\centering
\caption{Replicator-Mutator system, $N{=}5$ (\textbf{Q5}: qualitative behaviour). Mean $\pm$ std over 200 test initial conditions.}
\label{tab:replicator_results}
\footnotesize
\setlength{\tabcolsep}{4pt}
\begin{tabular}{@{}lcccc@{}}
\toprule
\textbf{Metric} & \textbf{Unconst.} & \textbf{Penalty} & \textbf{Ours} & \textbf{IF} \\
\midrule
MSE (Train) & $8.31\text{e-}5${\scriptsize $\pm 1.83\text{e-}4$} & $1.03\text{e-}2${\scriptsize $\pm 1.33\text{e-}2$} & $\mathbf{1.38\text{e-}5}${\scriptsize $\pm 3.62\text{e-}5$} & $6.0\times$ \\
MSE (Extrap) & $7.46\text{e-}5${\scriptsize $\pm 1.60\text{e-}4$} & $7.81\text{e-}3${\scriptsize $\pm 1.33\text{e-}2$} & $\mathbf{1.19\text{e-}5}${\scriptsize $\pm 2.75\text{e-}5$} & $6.3\times$ \\
MSE (Total) & $7.88\text{e-}5${\scriptsize $\pm 1.71\text{e-}4$} & $9.07\text{e-}3${\scriptsize $\pm 1.18\text{e-}2$} & $\mathbf{1.28\text{e-}5}${\scriptsize $\pm 3.18\text{e-}5$} & $6.2\times$ \\
\midrule
Viol.\ (Mean) & $3.08\text{e-}4${\scriptsize $\pm 6.54\text{e-}5$} & $3.11\text{e-}4${\scriptsize $\pm 1.75\text{e-}4$} & $\mathbf{2.79\text{e-}5}${\scriptsize $\pm 6.97\text{e-}6$} & $11\times$ \\
\bottomrule
\end{tabular}
\end{table}

Table~\ref{tab:replicator_results} shows that the skew-symmetric architecture achieves both $\mathbf{6\times}$ lower trajectory MSE and $\mathbf{11\times}$ lower simplex violation than the best baseline. This combination is essential for qualitative fidelity: maintaining simplex membership ($\sum x_i = 1$, $x_i \geq 0$) ensures that predicted species frequencies remain biologically meaningful, while lower MSE ensures the correct dynamical regime is reproduced. The penalty method fails dramatically: competing with the data-fitting loss degrades MSE by nearly three orders of magnitude while achieving negligible violation improvement.

\begin{figure}[htbp]
  \centering
  \includegraphics[width=0.75\textwidth]{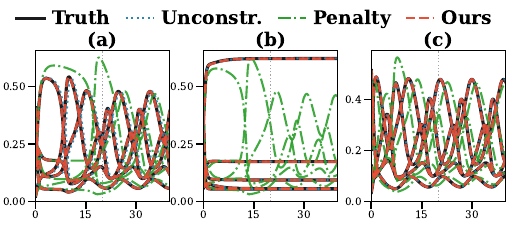}
  \caption{Replicator-Mutator system (\textbf{Q5}). Trajectories for three held-out initial conditions exhibiting qualitatively different dynamics: (a) transient oscillation, (b) monotone convergence, and (c) sustained oscillations. The vertical dashed line marks the training horizon.}
  \label{fig:q5_replicator}
\end{figure}

\subsection{Q6: Scalability}
\label{sec:q6}

We test whether compiled constraints remain effective as structural complexity grows, using the extended pendulum---a system that requires simultaneous preservation of a Hamiltonian and a Casimir invariant in a state-dependent Poisson structure. Training uses a single trajectory over $t \in [0, 30]$. Full results appear in Table~\ref{tab:results_pendulum} (Appendix~\ref{app:q6_results}).

Enforcing two invariants simultaneously does not degrade performance -- in contrast, the Poisson INN achieves $\mathbf{40\times}$ lower extrapolation MSE than the Neural ODE and \textbf{four orders of magnitude} lower than the penalty baseline at $10\times$ extrapolation. Both Casimir and Hamiltonian deviations remain tightly bounded, confirming that compiled constraints scale gracefully when multiple invariants must be enforced simultaneously. Figure~\ref{fig:3d_phase_space}(b) shows the Poisson INN tracking the true trajectory through $10\times$ extrapolation in the $(u, v, r)$ coordinate space.

\subsection{Q7: Composability}
\label{sec:q7}

Finally, we test whether invariants of heterogeneous type---some analytically known, others learned from data---can be composed within a single compiled architecture, using the equal-mass two-body gravitational problem with four simultaneous conserved quantities. We train on 1000 trajectories (500 steps per trajectory) and evaluate on 200 held-out trajectories over $4\times$ the training horizon.

Following Section~\ref{sec:first_integral}, we enforce all four invariants through manifold projection. The momentum constraints $p_x(u) = \dot{x}_1 + \dot{x}_2$ and $p_y(u) = \dot{y}_1 + \dot{y}_2$ are embedded as known analytical gradients, while two additional conserved quantities are learned: a Hamiltonian network $H_\phi(u)$ and a general first-integral network $V_\psi(u)$.

\begin{table}[htbp]
\centering
\caption{Two-body gravitational system (\textbf{Q7}: composability). Mean $\pm$ std over 200 test trajectories. IF relative to best baseline.}
\label{tab:composability}
\footnotesize
\setlength{\tabcolsep}{4pt}
\begin{tabular}{@{}lcccc@{}}
\toprule
\textbf{Metric} & \textbf{Unconst.} & \textbf{PINNs} & \textbf{Ours} & \textbf{IF} \\
\midrule
MSE (Total) & 2.01e-05{\scriptsize $\pm$ 1.79e-05} & 6.75e-05{\scriptsize $\pm$ 9.91e-05} & \textbf{2.92e-06}{\scriptsize $\pm$ 1.62e-06} & $\mathbf{6.9\times}$ \\
\midrule
Energy Dev. & 2.94e-04{\scriptsize $\pm$ 2.49e-04} & 4.84e-04{\scriptsize $\pm$ 3.04e-04} & \textbf{1.81e-04}{\scriptsize $\pm$ 9.11e-05} & $\mathbf{1.6\times}$ \\
Momentum Dev. & 1.77e-04{\scriptsize $\pm$ 6.89e-05} & 5.18e-04{\scriptsize $\pm$ 2.51e-04} & \textbf{2.42e-07}{\scriptsize $\pm$ 1.16e-07} & $\mathbf{731\times}$ \\
Ang-Mom Dev. & 4.86e-04{\scriptsize $\pm$ 4.61e-04} & 7.07e-04{\scriptsize $\pm$ 4.95e-04} & \textbf{2.75e-04}{\scriptsize $\pm$ 1.78e-04} & $\mathbf{1.8\times}$ \\
\bottomrule
\end{tabular}
\end{table}

Table~\ref{tab:composability} demonstrates that heterogeneous invariant types compose effectively. The analytically embedded momentum constraints yield the most dramatic result: momentum deviation drops to ${\sim}10^{-7}$, a $\mathbf{731\times}$ improvement over the best baseline. Crucially, embedding the known constraints does not impede learning of unknown ones---energy and angular momentum deviations are also the lowest among all methods, and total trajectory MSE improves by $\mathbf{6.9\times}$ over the unconstrained Neural ODE. The PINNs-style baseline, despite penalising all four known invariants, actually performs \emph{worse} than the unconstrained model on MSE and several deviation metrics, likely because four simultaneous penalty terms create conflicting gradient signals during optimization.

\section{Conclusion}

We have proposed the Invariant Compiler framework, a vision for systematically building Neural ODEs that satisfy physical constraints by construction. Unlike penalty-based methods that merely encourage constraint satisfaction, the compiler paradigm \emph{rewrites} learned dynamics into a form where invariants hold exactly throughout simulation and training. Future work will expand the compiler to handle inequality and contact constraints, discrete events and regime switching, and integration with differentiable simulators\cite{suh2022differentiablesimulatorsbetterpolicy}. We also plan to investigate automatic invariant discovery from data, extend the framework to PDEs \cite{arora2024invariantphysicsinformedneuralnetworks} and stochastic systems, and improve scalability to high-dimensional\cite{zang2020neural} settings. Applications in robotics \cite{howell2025dojodifferentiablephysicsengine}and molecular dynamics remain a key motivation. We hope this framework will serve as a foundation for building trustworthy scientific machine learning models.

\newpage
\bibliography{main}


\newpage
\appendix

\setcounter{table}{0}
\renewcommand{\thetable}{A\arabic{table}}

\setcounter{figure}{0}
\renewcommand{\thefigure}{A\arabic{figure}}

\section{Simplex Preservation: Full Derivation}
\label{app:simplex}

\subsection{Mathematical Formulation}

The probability simplex is defined as:
\begin{align}
\Delta^{n-1} = \left\{ x \in \mathbb{R}^n : \sum_{i=1}^n x_i = 1, \; x_i \geq 0 \right\}.
\end{align}
This encodes two requirements simultaneously: (i) a \textbf{sum constraint} $\sum_i x_i = 1$, and (ii) \textbf{non-negativity} $x_i \geq 0$.

\subsection{Motivation}

Many real-world systems evolve on probability simplices, where state variables represent proportions or frequencies that must sum to a fixed total while remaining non-negative. In compartmental epidemiology (SIR, SEIR, and variants) \cite{kermack1927contribution}, the population is partitioned into mutually exclusive categories whose fractions must sum to one---and negative populations are physically meaningless. In market share dynamics, competing firms' shares $s_i(t) \geq 0$ must satisfy $\sum_i s_i(t) = 1$ \cite{kimball1957industrial}. In population genetics, allele frequencies form a probability distribution that cannot create or destroy probability mass \cite{wright1931mendelian}.

Quantum mechanics provides inspiration for preserving such constraints. The state evolves via unitary operators that preserve the norm $\|U(t)\|_2^2 = 1$ \cite{NielsenChuang2010}. This norm preservation arises from skew-Hermitian generators: if $\frac{dU}{dt} = AU$ with $A^\dagger = -A$, then $\frac{d}{dt}(U^\dagger U) = 0$. We exploit an analogous geometric structure to preserve the simplex.

\subsection{Detailed Construction}

\subsubsection{Step 1: Square-Root Embedding}

Define the ``amplitude'' vector $u \in \mathbb{R}^n$ via:
\begin{align}
u_i = \sqrt{x_i}, \quad \text{so that} \quad \|u\|_2^2 = \sum_i u_i^2 = \sum_i x_i = 1.
\end{align}
This maps the simplex $\Delta^{n-1}$ into the unit sphere $S^{n-1} = \{u : \|u\|_2 = 1\}$.

\subsubsection{Step 2: Norm-Preserving Dynamics on the Sphere}

We evolve $u(t)$ via skew-symmetric dynamics:
\begin{align}
\frac{du}{dt} = A(u) \, u, \quad \text{where } A^\top = -A.
\end{align}
Such dynamics generate rotations on the sphere. At each instant, $A(u) \, u$ is orthogonal to $u$, producing infinitesimal rotations that preserve $\|u\|_2 = 1$.

For a Neural ODE parameterization, let $f_\theta(u)$ be an $n \times n$ matrix-valued network output. We construct the skew-symmetric dynamics matrix as:
\begin{align}
A(u) = \frac{1}{2} \left( f_\theta(u) - f_\theta(u)^\top \right).
\end{align}

The rotational dynamics can carry the state $u(t)$ outside the positive orthant, yielding negative components $u_i < 0$. This is not only permissible but essential for expressive dynamics.

\subsubsection{Step 3: Recovery via Squaring}

The physical simplex coordinates are recovered by squaring:
\begin{align}
x_i(t) = u_i(t)^2.
\end{align}
This map automatically guarantees \textbf{non-negativity}: regardless of the sign of $u_i$, we have $x_i = u_i^2 \geq 0$. Combined with norm preservation ($\sum_i u_i^2 = 1$), we obtain:
\begin{align}
\sum_i x_i(t) = \sum_i u_i(t)^2 = \|u(t)\|_2^2 = 1.
\end{align}

For the SIR model, the physical compartment values are \cite{beneduci2021unifying,leon2007quantum}:
\begin{align}
\begin{pmatrix} S(t) \\ I(t) \\ R(t) \end{pmatrix} = N \begin{pmatrix} u_1^2(t) \\ u_2^2(t) \\ u_3^2(t) \end{pmatrix}.
\end{align}

\subsection{Verification of Simplex Preservation}

\textbf{Sum constraint.} We compute:
\begin{align}
\frac{d}{dt} \|u\|_2^2 &= \frac{d}{dt} (u^\top u) = 2 u^\top \frac{du}{dt} = 2 u^\top A u.
\end{align}
Since $A$ is skew-symmetric, $u^\top A u = 0$ for any vector $u$. Thus $\|u(t)\|_2^2 = 1$ for all $t$, which implies $\sum_i x_i(t) = 1$.

\textbf{Non-negativity.} The squaring map $x_i = u_i^2$ guarantees $x_i \geq 0$ by construction, independent of the sign of $u_i$.

Together, these ensure that $x(t) \in \Delta^{n-1}$ for all $t \geq 0$. The simplex constraint is \emph{automatically guaranteed} by the geometry---no post-hoc projection or soft penalty is needed \cite{murari2022structure}. In this view, learning epidemic dynamics becomes learning a vector field on the $(n-1)$-sphere that, when composed with the squaring map, respects the simplex geometry \cite{bronstein2021geometricdl}.

\section{Lorentz Cone: Full Derivation}
\label{app:lorentz}

\subsection{Mathematical Formulation}

The Lorentz cone (also called the second-order cone or ice-cream cone) is defined as:
\begin{equation}
    \mathcal{L}^{n+1} = \left\{ (t, x) \in \mathbb{R} \times \mathbb{R}^n : t \geq \|x\|_2 \right\}.
\end{equation}

\subsection{Motivation}

In Einstein's theory of special relativity, causal influences cannot propagate faster than the speed of light $c$ \cite{TaylorWheeler1992SpacetimePhysics}. For an event at spacetime coordinates $(t, x)$ to lie within the causal future of the origin, the time coordinate must satisfy $t \geq \|x\|_2 / c$. In natural units where $c = 1$, this constraint defines the \emph{future light cone}. Beyond relativistic physics, the Lorentz cone appears in robust optimization, signal processing, and any setting where one variable bounds the Euclidean norm of others \cite{Boyd2004Convex}.

\subsection{Detailed Construction}

To ensure that learned trajectories remain within a cone $K$, we require the vector field to satisfy a \textbf{viability condition}: at every point, the dynamics must not point outward from the cone \cite{Aubin1991Viability}. At any point $z \in K$, the \emph{tangent cone} $T_K(z)$ consists of all directions along which one can move while remaining in $K$:
\begin{equation}
    T_K(z) = \left\{ v : \exists \epsilon > 0 \text{ such that } z + tv \in K \text{ for all } t \in [0, \epsilon] \right\}.
\end{equation}
For a system
\begin{equation}
    \frac{dz}{dt} = f(z),
\end{equation}
the viability condition requires $f(z) \in T_K(z)$ for all $z \in K$.

Given an arbitrary vector field $\hat{f}(z)$, we enforce the viability condition by projecting onto the tangent cone:
\begin{equation}
    \frac{dz}{dt} = \pi_{T_K(z)}(\hat{f}(z)).
\end{equation}

\subsubsection{Projection onto the Lorentz Cone}

Let $K = \mathcal{L}^{n+1}$, $z = (t, x) \in K$ and let $\hat f(z) = (a(z), b(z))$ be a vector field, where $a$ and $b$ are mappings $a: \mathbb{R}^{n+1} \to \mathbb{R}$ and $b: \mathbb{R}^{n+1} \to \mathbb{R}^n$. 

\paragraph{Case 1: Interior points.} When $t > \|x\|_2$, the tangent cone is the entire space $\mathbb{R}^{n+1}$, so $\pi_{T_K}(a, b) = (a, b)$.

\paragraph{Case 2: Boundary points.} When $t = \|x\|_2$ with $x \neq 0$, the outward normal is $\nabla \phi = (1, -u)$ where $u = x/\|x\|_2$. The tangent cone is $T_K(z) = \{(a,b) : a \geq u^\top b\}$. If $a \geq u^\top b$, no projection is needed. Otherwise:
\begin{equation}
    \pi_{T_K}(a, b) = (a, b) - \frac{a - u^\top b}{2} (1, -u).
\end{equation}

\paragraph{Case 3: The apex.} When $z = 0$, the tangent cone is $\mathcal{L}^{n+1}$ itself. Let $\beta = \|b\|_2$. If $\beta \leq a$: $(a, b)$ is already in the cone. If $\beta \leq -a$: $\pi_{T_K}(a, b) = 0$. Otherwise:
\begin{equation}
    \pi_{T_K}(a, b) = \left( \frac{a + \beta}{2}, \frac{a + \beta}{2\beta} b \right).
\end{equation}

\subsection{Verification}

\textbf{Interior points.} No correction is needed; any trajectory starting in the interior either remains or reaches the boundary.

\textbf{Boundary points.} After projection, $a' \geq u^\top b'$, so $\nabla \phi^\top \dot{z} = a' - u^\top b' \geq 0$---the trajectory cannot exit through the boundary.

\textbf{The apex.} The projection maps into $\mathcal{L}^{n+1}$ itself, so $\dot{z} \in \mathcal{L}^{n+1}$ and any trajectory from the apex moves into the cone.

\section{PSD Cone: Full Derivation}
\label{app:psd}

\subsection{Mathematical Formulation}

The positive semidefinite cone is:
\begin{equation}
    \mathcal{S}_{+}^n = \left\{ X \in \mathbb{R}^{n \times n} : X = X^\top, \; X \succeq 0 \right\}.
\end{equation}

\subsection{Motivation}

In filtering, control, and uncertainty quantification \cite{Anderson2005OptimalFiltering}, the state often includes a covariance matrix $P = \mathbb{E}[(X-\mu)(X-\mu)^\top]$ that must remain positive semidefinite. The Kalman filter propagates $P$ via a matrix Riccati equation; numerical errors can cause $P$ to lose positive semidefiniteness, leading to catastrophic filter divergence.

\subsection{Construction}

Any $P \in \mathcal{S}_{+}^n$ admits a factorization $P = LL^\top$, where $L$ is lower triangular with non-negative diagonal entries \cite{Bierman1977Factorization}. Let $g_\theta : \mathbb{R}^{n(n+1)/2} \to \mathbb{R}^{n(n+1)/2}$ be a neural network \cite{Ko2025CholeskyKalmanNet}:
\begin{equation}
    \frac{dL_{ij}}{dt} = g_\theta(L)_{ij}, \quad i \geq j, \qquad P(t) = L(t) L(t)^\top.
\end{equation}

\subsection{Verification}

\textbf{Symmetry.} $(LL^\top)^\top = LL^\top$.

\textbf{Positive semidefinite.} For all $v \in \mathbb{R}^n$, $v^\top P v = v^\top LL^\top v = \|L^\top v\|_2^2 \geq 0$.

This mirrors the structure of covariance matrices: $a^\top P a = \mathbb{E}[((X-\mu)^\top a)^2] \geq 0$.

\section{Center of Mass Frame: Full Derivation}
\label{app:com}

\subsection{Mathematical Formulation}

For $n$ interacting particles with masses $m_i$, positions $\mathbf{r}_i$, and velocities $\mathbf{v}_i$, the center of mass frame requires:
\begin{align}
\sum_{i=1}^{n} m_i \mathbf{r}_i = \mathbf{0}, \qquad \sum_{i=1}^{n} m_i \mathbf{v}_i = \mathbf{0}.
\end{align}

\subsection{Motivation}

For isolated systems of interacting particles---colliding billiard balls, planetary systems, molecular dynamics---the center of mass frame simplifies analysis by focusing on relative motion \cite{PrantlUKT22,BattagliaPLRK16,BishnoiJRK24}. A naive neural network predicting accelerations has no mechanism to guarantee $\sum_i m_i \mathbf{a}_i = \mathbf{0}$, causing drift from the center of mass frame.

\subsection{Construction}

Let $\bar{\mathbf{v}} = \frac{\sum_k m_k \mathbf{v}_k}{\sum_k m_k}$ and $\bar{\mathbf{a}} = \frac{\sum_k m_k \mathbf{a}_k}{\sum_k m_k}$. The constraint-preserving dynamics are:
\begin{align}
\frac{d\mathbf{r}_i}{dt} = \mathbf{v}_i - \bar{\mathbf{v}}, \qquad \frac{d\mathbf{v}_i}{dt} = \mathbf{a}_i - \bar{\mathbf{a}}.
\end{align}

\subsection{Verification}

\textbf{Position constraint.}
$\frac{d}{dt} \sum_i m_i \mathbf{r}_i = \sum_i m_i (\mathbf{v}_i - \bar{\mathbf{v}}) = \mathbf{0}$.

\textbf{Momentum constraint.}
$\frac{d}{dt} \sum_i m_i \mathbf{v}_i = \sum_i m_i (\mathbf{a}_i - \bar{\mathbf{a}}) = \mathbf{0}$.

\section{Stoichiometric Constraints: Full Derivation}
\label{app:stoich}

\subsection{Mathematical Formulation}

The state $c(t) \in \mathbb{R}^n$ (species concentrations) must satisfy:
\begin{align}
\mathbf{M}  c(t) = \mathbf{M}  c(0) \quad \text{for all } t, \qquad \text{equivalently,} \quad \mathbf{M} \frac{dc}{dt} = 0.
\end{align}

\subsection{Motivation}

Chemical and biological systems obey conservation laws: atoms are rearranged but never created or destroyed \cite{Feinberg2019}. The reaction $2\mathrm{H}_2 + \mathrm{O}_2 \to 2\mathrm{H}_2\mathrm{O}$ preserves hydrogen and oxygen counts. Analogous constraints arise in metabolic networks (conserved moieties) and ecological models (nutrient pools).

\subsection{Detailed Construction}

\subsubsection{The Molecular Matrix}

Following \citet{advanced}, $M$ has rows for conserved quantities, columns for species, with $m_{ij}$ the count of quantity $i$ in species $j$.

\textbf{Example: Water Formation.} Species $\mathrm{H}_2$, $\mathrm{O}_2$, $\mathrm{H}_2\mathrm{O}$; conserved elements H, O:
\begin{equation}
\mathbf{M} = \begin{pmatrix} 2 & 0 & 2 \\ 0 & 2 & 1 \end{pmatrix}.
\end{equation}

\subsubsection{The Balanced Transformation Matrix}

The null space basis $B$ of $M$ encodes all stoichiometrically valid directions. For water formation:
\begin{equation}
b_1 = (-2, -1, 2)^\top, \quad B = (-2, -1, 2)^\top,
\end{equation}
encoding $2\mathrm{H}_2 + \mathrm{O}_2 \to 2\mathrm{H}_2\mathrm{O}$.

\subsubsection{Null Space Projection}

$\frac{dc}{dt} = B r_\theta(c,t)$, where $r_\theta : \mathbb{R}^n \times \mathbb{R} \to \mathbb{R}^k$ and $k = \dim(\mathrm{Null}(\mathbf{M}))$.

\subsection{Verification}

Since $\mathbf{M}B = 0$: $\mathbf{M}  \frac{dc}{dt} = \mathbf{M}  B  r_\theta(c,t) = 0$.

\subsection{Interpretability}

This structure enables discovery of effective stoichiometry, reaction kinetics, and pathway dynamics from data, while guaranteeing physical validity.

\section{Hamiltonian / Poisson Structure: Full Derivation}
\label{app:poisson}

\subsection{Mathematical Formulation}

The system $\frac{du}{dt} = J(u) \nabla_u H(u)$ must preserve: (i) energy conservation $\frac{dH}{dt} = 0$, and (ii) the Jacobi identity for the Poisson bracket $\{F, G\} = (\nabla F)^\top J (\nabla G)$:
\begin{equation}
    \{F, \{G, K\}\} + \{G, \{K, F\}\} + \{K, \{F, G\}\} = 0.
\end{equation}

\subsection{Motivation}

The simplest case is canonical Hamiltonian mechanics \cite{Goldstein2002ClassicalMechanics,Greydanus2019HNN}:
\begin{equation}
    \frac{dq_i}{dt} = \frac{\partial H}{\partial p_i}, \qquad
    \frac{dp_i}{dt} = -\frac{\partial H}{\partial q_i},
\end{equation}
where the constant canonical matrix $J_0$ trivially satisfies the Jacobi identity. For general coordinates, enforcing the Jacobi identity for a state-dependent $J(u)$ is nontrivial.

\subsection{Detailed Construction}

\subsubsection{Latent Space Embedding}

Following \citet{jin2023learning}, an invertible neural network maps $z = g_\theta(u) = (q, p, c)^\top$ where $(q,p) \in \mathbb{R}^{2d}$ are canonical conjugate coordinates and $c \in \mathbb{R}^k$ are Casimir coordinates ($k = n - 2d$).

\subsubsection{Canonical Dynamics}

\begin{equation}
    \frac{dz}{dt} = J_0 \nabla_z K(z), \quad J_0 = \begin{pmatrix} 0 & I_d & 0 \\ -I_d & 0 & 0 \\ 0 & 0 & 0 \end{pmatrix},
\end{equation}
yielding $\dot{q} = \partial K/\partial p$, $\dot{p} = -\partial K/\partial q$, $\dot{c} = 0$. Physical coordinates recovered via $u' = g_\theta^{-1}(z')$.

\subsection{Verification}

\textbf{Energy conservation.} $\frac{dK}{dt} = (\nabla_z K)^\top J_0 (\nabla_z K) = 0$ by skew-symmetry.

\textbf{Jacobi identity.} The Poisson bracket is invariant under coordinate transformations. Since
\begin{equation}
    J(u) = \left(\frac{\partial z}{\partial u}\right)^{-1} J_0 \left(\frac{\partial z}{\partial u}\right)^{-\top}
\end{equation}
is the standard transformation law for Poisson tensors, and the Jacobi identity is preserved under smooth coordinate transformations, $J(u)$ automatically satisfies the Jacobi identity.

\section{Port-Hamiltonian Structure: Full Derivation}
\label{app:port_ham}

\subsection{Mathematical Formulation}

The system $\frac{du}{dt} = [J(u) - R(u)] \nabla_u H(u)$ must satisfy $\frac{dH}{dt} \leq 0$, where $J^\top = -J$ and $R \succeq 0$.

\subsection{Motivation}

Systems with friction, damping, or irreversible losses \cite{Duindam2009PortHamiltonianBook}. A pendulum with air resistance, an electrical circuit with resistance, a vibrating string with internal damping. Stability is built into the architecture by incorporating a dissipative term.

\subsection{Detailed Construction}

Following \citet{cheng2024learning}, in latent space:
\begin{equation}
    \frac{dz}{dt} = [J_0 - R_z(z)] \nabla_z K(z), \quad R_z(z) = L(z)L(z)^\top \succeq 0.
\end{equation}

\subsection{Verification}

\begin{align}
    \frac{dK}{dt} &= (\nabla_z K)^\top [J_0 - R_z] \nabla_z K = \underbrace{(\nabla_z K)^\top J_0 (\nabla_z K)}_{=0} - (\nabla_z K)^\top R_z (\nabla_z K) \leq 0.
\end{align}

\textbf{Convergence to equilibrium.} For Lyapunov stability with guaranteed convergence, parameterize $K(z)$ as convex with a unique minimizer \cite{Garg2021FixedTimeGradientFlows}.

\section{GENERIC / Thermodynamic Structure: Full Derivation}
\label{app:generic}

\subsection{Mathematical Formulation}

The GENERIC system $\frac{du}{dt} = J(u)\nabla H + M(u)\nabla S$ must satisfy: (i) $\frac{dH}{dt} = 0$, (ii) $\frac{dS}{dt} \geq 0$, and (iii) degeneracy conditions $J\nabla S = 0$ and $M\nabla H = 0$ \cite{Ottinger2005BeyondEquilibriumThermodynamics}.

\subsection{Motivation}

Thermodynamic systems exhibit both reversible (energy-conserving) and irreversible (entropy-producing) dynamics. Pure Hamiltonian mechanics is time-reversible and entropy-neutral; pure gradient flows dissipate energy. Real thermodynamic systems require both structures operating independently \cite{Zhang2022GFINNs}.

\subsection{Detailed Construction}

\subsubsection{Casimir-Dependent Entropy}

$S(z) = \tilde{S}(c)$, so $\nabla_z S = (0, 0, \nabla_c \tilde{S})^\top$. Then:
\begin{equation}
    J_0 \nabla_z S = \begin{pmatrix} 0 & I_d & 0 \\ -I_d & 0 & 0 \\ 0 & 0 & 0 \end{pmatrix} \begin{pmatrix} 0 \\ 0 \\ \nabla_c \tilde{S} \end{pmatrix} = \mathbf{0}.
\end{equation}

\subsubsection{Projected Dissipation Matrix}

$M_z(z) = P_K \widehat{M}_z P_K$ where $P_K = I - \frac{\nabla K (\nabla K)^\top}{\|\nabla K\|^2 + \varepsilon}$ and $\widehat{M}_z = LL^\top \succeq 0$.

\subsubsection{Complete Dynamics}

$\frac{dz}{dt} = J_0 \nabla_z K(z) + M_z(z) \nabla_z S(z)$ \cite{Sipka2021LearningGENERIC}.

\subsection{Verification}

\textbf{Degeneracy conditions.} (1) $J_0 \nabla_z S = 0$: automatic from Casimir parameterization. (2) $M_z \nabla_z K = P_K \widehat{M}_z P_K \nabla_z K = 0$ since $P_K \nabla_z K = 0$.

\textbf{Properties of $M_z$.} Symmetry: $(P_K \widehat{M}_z P_K)^\top = P_K \widehat{M}_z P_K$. PSD: $y^\top M_z y = (P_K y)^\top \widehat{M}_z (P_K y) \geq 0$.

\textbf{Energy conservation.} $\frac{dK}{dt} = (\nabla K)^\top J_0 \nabla K + (\nabla K)^\top M_z \nabla S = 0 + 0 = 0$.

\textbf{Entropy production.} $\frac{dS}{dt} = (\nabla S)^\top J_0 \nabla K + (\nabla S)^\top M_z \nabla S = 0 + (\nabla S)^\top M_z \nabla S \geq 0$.

\section{First Integral Preservation: Full Derivation}
\label{app:first_integral}

\subsection{Mathematical Formulation}

The state $u(t)$ is required to satisfy $V(u(t)) = V(u_0)$ for learned conserved quantities $V = (V_1, \ldots, V_m)^\top$.

\subsection{Motivation}

Many dynamical systems harbor hidden conserved quantities whose functional form is not given \emph{a priori}. The challenge is twofold: we do not know the functional form of $V$, and may not know how many independent conserved quantities exist.

\subsection{Detailed Construction}

\subsubsection{Learning the Constraint Manifold}

Following \citet{FINDE2023}, we parameterize each conserved quantity as a neural network $V_i(u; \phi_i)$. The Jacobian $\nabla V(u)$ is $m \times n$.

\subsubsection{Tangent Space Projection}

For a single constraint, the projection removes the normal component:
\begin{align}
\frac{du}{dt} = \hat{f}(u) - \frac{\nabla V(u)^\top \hat{f}(u)}{\|\nabla V(u)\|^2} \nabla V.
\end{align}

Following \citet{PNDE2024}, for multiple constraints, we adopt the Moore--Penrose pseudo-inverse \cite{Hauswirth2020ObliqueProjectedDynamicalSystems}:
\begin{align}
P = \nabla V^\top (\nabla V\nabla V^\top)^+ \nabla V, \qquad \frac{du}{dt} = (I - P(u))\hat{f}(u).
\end{align}

The pseudo-inverse handles redundant constraints gracefully: if the network discovers $V_3 = 2V_1 + \text{const}$, the projection is unchanged since $\operatorname{span}\{\nabla V_1, \nabla V_2, \nabla V_3\} = \operatorname{span}\{\nabla V_1, \nabla V_2\}$.

\subsection{Verification}

$P$ is the orthogonal projector onto $\operatorname{row}(\nabla V)$. Since $\nabla V_i^\top$ is a row of $\nabla V$ and $(I-P)$ projects onto $\operatorname{null}(\nabla V)$:
\begin{align}
\frac{d}{dt} V_i(u) = \nabla V_i^\top (I-P)\hat{f} = 0 \quad \text{for all } i.
\end{align}

\textbf{Remark.} The learned conserved quantities are not guaranteed to mutually commute under the Poisson bracket, which may be relevant for Hamiltonian systems where integrability structure is important.

\section{ODE System Specifications}
\label{app:ode_systems}

This appendix provides detailed specifications for all eleven dynamical systems introduced in Section~\ref{sec:experimental_setup}. The systems are organised into three categories based on their invariant structure.

\subsection{Geometric and Algebraic Constraints}

These systems possess known algebraic invariant sets---simplices, stoichiometric subspaces, or cones---that must be preserved exactly.

\paragraph{(i)~SIR epidemiological model ($\mathbb{R}^3$).}
The SIR model partitions a population into Susceptible ($S$), Infected ($I$), and Recovered ($R$) compartments with infection rate $\beta = 0.4$ and recovery rate $\gamma = 0.1$:
\begin{equation}
    \dot{S} = -\beta S I, \qquad \dot{I} = \beta S I - \gamma I, \qquad \dot{R} = \gamma I,
\end{equation}
with the simplex constraint $S(t) + I(t) + R(t) = 1$ for all $t$. Following Section~\ref{sec:simplex}, we embed the simplex into the unit sphere via the square-root map and employ skew-symmetric dynamics.

\paragraph{(ii)~Chemical reaction network ($\mathbb{R}^6$).}
Six species (CO, H\textsubscript{2}O, CO\textsubscript{2}, H\textsubscript{2}, O\textsubscript{2}, CH\textsubscript{4}) undergo the water--gas shift reaction (CO + H\textsubscript{2}O $\rightleftharpoons$ CO\textsubscript{2} + H\textsubscript{2}), combustion ($2\text{CO} + \text{O}_2 \rightleftharpoons 2\text{CO}_2$), and an inactive steam methane reforming channel (CH\textsubscript{4} + H\textsubscript{2}O $\rightleftharpoons$ CO + 3H\textsubscript{2}). The stoichiometric matrix $\mathbf{S} \in \mathbb{R}^{3 \times 6}$ encodes conservation of carbon, hydrogen, and oxygen atoms. Following Section~\ref{sec:stoich}, we parameterise dynamics via the null-space basis, ensuring $\mathbf{S} \cdot \dot{\mathbf{c}} = \mathbf{0}$ by construction.

\paragraph{(iii)~NOx reaction network ($\mathbb{R}^5$).}
Five species (NO, O\textsubscript{2}, NO\textsubscript{2}, N\textsubscript{2}O\textsubscript{4}, N\textsubscript{2}O\textsubscript{3}) participate in three coupled reactions with strongly nonlinear kinetics:
\begin{align}
    \text{R1:}\;&\; 2\text{NO} + \text{O}_2 \rightleftharpoons 2\text{NO}_2, \notag \\
    \text{R2:}\;&\; 2\text{NO}_2 \rightleftharpoons \text{N}_2\text{O}_4, \notag \\
    \text{R3:}\;&\; \text{NO} + \text{NO}_2 \rightleftharpoons \text{N}_2\text{O}_3.
\end{align}
The reaction rates feature product inhibition ($\text{R1}$: $(1 + \alpha\, c_{\text{NO}_2})^{-2}$ and exponential back-reaction), substrate inhibition ($\text{R2}$: $(1 + (c_{\text{NO}_2}/K_2)^2)^{-1}$), and fractional-power kinetics ($\text{R3}$: $c_{\text{N}_2\text{O}_3}^{0.8}$). Two elements (N, O) are conserved via the stoichiometric matrix $\mathbf{S} \in \mathbb{R}^{2 \times 5}$.

\paragraph{(iv)~Lorentz cone spiral ($\mathbb{R}^3$).}
An expanding spiral on the boundary of the Lorentz cone $\mathcal{L}^3 = \{(t, x_1, x_2) : t \geq \|(x_1, x_2)\|_2\}$, with growth rate $\alpha{=}0.08$ and angular velocity $\omega{=}0.4$.

\paragraph{(v)~Coupled radial-angular dynamics ($\mathbb{R}^3$).}
Nonlinear dynamics on the Lorentz cone with coupled radial and angular evolution:
\begin{equation}
    \frac{dr}{d\tau} = \alpha \, r\!\left(1 - \frac{r}{K}\right) + \beta \, r \cos(n\theta), \qquad
    \frac{d\theta}{d\tau} = \omega + \frac{\gamma}{r^2},
\end{equation}
with $\alpha = 1.0$, $K = 5.0$, $\beta = 0.8$, $\omega = 1.0$, and $\gamma = 0.5$. Logistic radial growth is modulated by angular position while angular velocity depends inversely on radius, creating rich spiral trajectories.

\paragraph{(vi)~Replicator-mutator system ($\mathbb{R}^5$).}
An $N{=}5$ species evolutionary dynamics model~\cite{PaisLeonard2011LimitCycles}:
\begin{equation}
    \dot{x}_i = \sum_{j=1}^{N} Q_{ji}\, x_j\, (\mathbf{B}\mathbf{x})_j \;-\; \bar{\phi}(\mathbf{x})\, x_i, \qquad i = 1,\dots,N,
\end{equation}
where $\mathbf{B} \in \mathbb{R}^{N \times N}$ is a payoff matrix, $Q$ is a mutation matrix with rate $\mu = 0.15$, $\bar{\phi}(\mathbf{x}) = \mathbf{x}^\top \mathbf{B}\mathbf{x}$ is the mean fitness, and a speed-up factor $\alpha = 10$ accelerates the dynamics. The system preserves the simplex constraint $\sum_i x_i = 1$ and, depending on initial conditions, exhibits sustained oscillations, monotone convergence, or transient oscillatory behaviour followed by equilibration.

\subsection{Energetic and Thermodynamic Structure}

These systems possess conserved or dissipated quantities whose structure must be learned from data.

\paragraph{(vii)~Lotka-Volterra predator-prey ($\mathbb{R}^2$).}
The predator-prey model with parameters $\alpha = 1.0$, $\beta = \delta = \gamma_{\mathrm{LV}} = 0.5$:
\begin{equation}
    \dot{x} = \alpha x - \beta xy, \qquad \dot{y} = \delta xy - \gamma_{\mathrm{LV}} y,
\end{equation}
exhibits Hamiltonian structure with a non-canonical Poisson bracket. The Poisson INN learns a coordinate transformation to canonical variables where dynamics take standard Hamiltonian form.

\paragraph{(viii)~Damped harmonic oscillator ($\mathbb{R}^2$).}
With damping coefficient $\gamma = 0.15$ and natural frequency $\omega = 1.0$, the system satisfies $\dot{E} = -\gamma \dot{q}^2 \leq 0$. The Port-Hamiltonian INN learns a Hamiltonian $K(z)$ and a state-dependent dissipation matrix $R(z) = L(z)L(z)^\top \succeq 0$.

\paragraph{(ix)~Thermomechanical system ($\mathbb{R}^3$).}
A damped oscillator coupled to a heat bath~\cite{Oettinger2018GENERICIntegrators}:
\begin{align}
    \dot{q} &= p, \qquad \dot{p} = -\omega^2 q - \gamma p, \qquad \dot{\theta} = \frac{\gamma p^2}{C_v},
\end{align}
with $\gamma = 0.15$, $\omega = 1.0$, and heat capacity $C_v = 1.0$. Total energy is conserved while entropy increases monotonically. The GENERIC INN learns separate energy and entropy functions with architecturally enforced degeneracy conditions.

\paragraph{(x)~Extended pendulum ($\mathbb{R}^3$).}
Starting from $H(p,q) = \tfrac{1}{2}p^2 - \cos q$, the phase space is augmented with a third coordinate $c$ and a nonlinear transformation $(u,v,r) = (p,\, q,\, p^2 + q^2 + c)$ yields a Poisson system with a state-dependent structure matrix~\cite{jin2023learning}:
\begin{equation}
    \begin{pmatrix} \dot{u} \\ \dot{v} \\ \dot{r} \end{pmatrix}
    = \underbrace{\begin{pmatrix} 0 & -1 & -2v \\ 1 & 0 & 2u \\ 2v & -2u & 0 \end{pmatrix}}_{B(u,v,r)}
    \nabla K(u,v,r),
\end{equation}
with $K(u,v,r) = \tfrac{1}{2}u^2 - \cos v + ur - u^3 - uv^2$. The system possesses two independent invariants: the Hamiltonian $K$ and the Casimir $C(u,v,r) = r - u^2 - v^2$. The Poisson INN learns canonical coordinates where both invariants are preserved by construction.

\subsection{Multi-Invariant Systems}

This category tests whether invariants of heterogeneous type---some analytically known, others learned from data---can be composed within a single compiled architecture.

\paragraph{(xi)~Two-body gravitational system ($\mathbb{R}^8$).}
The equal-mass two-body problem~\cite{FINDE2023} with state $u = (x_1, x_2, y_1, y_2, \dot{x}_1, \dot{x}_2, \dot{y}_1, \dot{y}_2)$ evolving under Newtonian gravity. The system possesses four independent conserved quantities: total energy $H$, linear momentum in $x$ and $y$ ($p_x = \dot{x}_1 + \dot{x}_2 = 0$, $p_y = \dot{y}_1 + \dot{y}_2 = 0$ in the centre-of-mass frame), and angular momentum $L = \sum_i (x_i \dot{y}_i - y_i \dot{x}_i)$. Within the manifold projection framework, two invariants (linear momentum) are analytically known and supplied as analytical gradients, while the remaining two (energy, angular momentum) must be discovered from data and enforced simultaneously.

\section{Supplementary Experimental Results}
\label{app:supplementary}

This appendix reports results for the supplementary systems introduced in Section~\ref{sec:experimental_setup}. 

\subsection{SIR Epidemiological Model (Q1 Supplementary)}

The SIR model is trained on 100 trajectories with different initial conditions and evaluated on 20 held-out initial conditions. Following Section~\ref{sec:simplex}, we embed the simplex into the unit sphere via the square-root map and employ skew-symmetric dynamics.

\begin{table}[H]
\centering
\caption{SIR model (Q1 supplementary). Mean $\pm$ std over 20 test initial conditions.}
\label{tab:sir_results}
\footnotesize
\setlength{\tabcolsep}{4pt}
\begin{tabular}{@{}lcccc@{}}
\toprule
\textbf{Metric} & \textbf{Unconst} & \textbf{Penalty} & \textbf{Ours} & \textbf{IF} \\
\midrule
MSE (Train) & $3.62\text{e-}3${\scriptsize $\pm 4.33\text{e-}3$} & $1.20\text{e-}2${\scriptsize $\pm 8.37\text{e-}3$} & $\mathbf{1.37\text{e-}3}${\scriptsize $\pm 3.83\text{e-}3$} & $2.6\times$ \\
MSE (Extrap) & $2.67\text{e}{+}2${\scriptsize $\pm 7.46\text{e}{+}2$} & $1.82\text{e-}2${\scriptsize $\pm 1.73\text{e-}2$} & $\mathbf{7.79\text{e-}3}${\scriptsize $\pm 2.22\text{e-}2$} & $2.3\times$ \\
MSE (Total) & $1.33\text{e}{+}2${\scriptsize $\pm 3.73\text{e}{+}2$} & $1.51\text{e-}2${\scriptsize $\pm 1.23\text{e-}2$} & $\mathbf{4.58\text{e-}3}${\scriptsize $\pm 1.29\text{e-}2$} & $3.3\times$ \\
\midrule
Viol.\ (Mean) & $4.52\text{e}{+}0${\scriptsize $\pm 7.98\text{e}{+}0$} & $1.59\text{e-}2${\scriptsize $\pm 1.13\text{e-}2$} & $\mathbf{1.69\text{e-}5}${\scriptsize $\pm 9.52\text{e-}6$} & $941\times$ \\
\bottomrule
\end{tabular}
\end{table}

\subsection{Chemical Reaction Network (Q1 Supplementary)}

The six-species chemical system is trained on 100 trajectories and evaluated on 20 held-out initial conditions. The null-space parameterisation ensures $\mathbf{S} \cdot \dot{\mathbf{c}} = \mathbf{0}$ by construction.

\begin{table}[H]
\centering
\caption{Chemical reaction network (Q1 supplementary). Mean $\pm$ std over 20 test initial conditions.}
\label{tab:chem_results}
\footnotesize
\setlength{\tabcolsep}{4pt}
\begin{tabular}{@{}lcccc@{}}
\toprule
\textbf{Metric} & \textbf{Unconstrained} & \textbf{Penalty} & \textbf{Null-Space (Ours)} & \textbf{IF} \\
\midrule
MSE (Train) & $6.24\text{e-}6${\scriptsize $\pm 6.33\text{e-}6$} & $7.52\text{e-}3${\scriptsize $\pm 7.94\text{e-}3$} & $\mathbf{4.43\text{e-}6}${\scriptsize $\pm 2.65\text{e-}6$} & $1.4\times$ \\
MSE (Extrap) & $2.89\text{e-}4${\scriptsize $\pm 4.48\text{e-}4$} & $3.82\text{e-}1${\scriptsize $\pm 1.01\text{e-}1$} & $\mathbf{1.26\text{e-}4}${\scriptsize $\pm 1.40\text{e-}4$} & $2.3\times$ \\
MSE (Total) & $1.47\text{e-}4${\scriptsize $\pm 2.26\text{e-}4$} & $1.95\text{e-}1${\scriptsize $\pm 5.09\text{e-}2$} & $\mathbf{6.51\text{e-}5}${\scriptsize $\pm 6.94\text{e-}5$} & $2.3\times$ \\
\midrule
Viol.\ (Mean) & $2.50\text{e-}2${\scriptsize $\pm 1.14\text{e-}2$} & $5.53\text{e-}3${\scriptsize $\pm 2.66\text{e-}3$} & $\mathbf{3.16\text{e-}6}${\scriptsize $\pm 1.09\text{e-}6$} & $1750\times$ \\
\bottomrule
\end{tabular}
\end{table}

\subsection{Lorentz Cone Spiral (Q2 Supplementary)}

The simple expanding spiral on the Lorentz cone boundary is trained on 100 trajectories and evaluated on 20 held-out initial conditions.

\begin{table}[H]
\centering
\caption{Lorentz cone spiral (Q2 supplementary). Mean $\pm$ std over 20 test initial conditions.}
\label{tab:lorentz_simple_results}
\footnotesize
\setlength{\tabcolsep}{4pt}
\begin{tabular}{@{}lcccc@{}}
\toprule
\textbf{Metric} & \textbf{Unconst.} & \textbf{Penalty} & \textbf{Ours} & \textbf{IF} \\
\midrule
MSE (Train) & $7.00\text{e-}8${\scriptsize $\pm 6.64\text{e-}8$} & $5.55\text{e-}5${\scriptsize $\pm 5.67\text{e-}5$} & $\mathbf{5.18\text{e-}8}${\scriptsize $\pm 6.17\text{e-}8$} & $1.4\times$ \\
MSE (Extrap) & $\mathbf{3.03\text{e-}2}${\scriptsize $\pm 3.83\text{e-}2$} & $1.48\text{e}{+}0${\scriptsize $\pm 1.92\text{e}{+}0$} & $7.05\text{e-}2${\scriptsize $\pm 9.98\text{e-}2$} & --- \\
MSE (Total) & $\mathbf{1.52\text{e-}2}${\scriptsize $\pm 1.91\text{e-}2$} & $7.40\text{e-}1${\scriptsize $\pm 9.62\text{e-}1$} & $3.53\text{e-}2${\scriptsize $\pm 4.99\text{e-}2$} & --- \\
\midrule
Viol.\ (Mean) & $3.79\text{e-}2${\scriptsize $\pm 2.65\text{e-}2$} & $9.27\text{e-}4${\scriptsize $\pm 1.15\text{e-}3$} & $\mathbf{2.85\text{e-}8}${\scriptsize $\pm 2.22\text{e-}8$} & $32\,500\times$ \\
\bottomrule
\end{tabular}
\end{table}

\subsection{Damped Harmonic Oscillator (Q3 Supplementary)}

The damped harmonic oscillator is trained on a single trajectory over $t \in [0, 30]$ and evaluated at $2\times$, $5\times$, and $10\times$ extrapolation.

\begin{table}[H]
\caption{Damped harmonic oscillator (Q3 supplementary).}
\label{tab:damped_results}
\centering
\footnotesize
\setlength{\tabcolsep}{3pt}
\begin{tabular}{@{}llcccc@{}}
\toprule
\textbf{Metric} & \textbf{Model} & \textbf{$\times 2$} & \textbf{$\times 5$} & \textbf{$\times 10$} & \textbf{IF ($\times 10$)} \\
\midrule
\multirow{3}{*}{MSE (Train)} & Unconst. & 2.81e-06 & 2.81e-06 & 2.81e-06 & \\
 & PINNs & 8.18e-06 & 8.18e-06 & 8.18e-06 & \\
 & Ours & \textbf{3.12e-07} & \textbf{3.12e-07} & \textbf{3.12e-07} & $9.0\times$ \\
\midrule
\multirow{3}{*}{MSE (Extrap)} & Unconst. & 8.65e-07 & 2.52e-07 & 1.30e-07 & \\
 & PINNs & 6.97e-06 & 1.95e-06 & 9.35e-07 & \\
 & Ours & \textbf{2.50e-07} & \textbf{1.45e-07} & \textbf{1.25e-07} & $1.04\times$ \\
\midrule
\multirow{3}{*}{MSE (Total)} & Unconst. & 1.84e-06 & 7.63e-07 & 3.97e-07 & \\
 & PINNs & 7.57e-06 & 3.19e-06 & 1.66e-06 & \\
 & Ours & \textbf{2.81e-07} & \textbf{1.78e-07} & \textbf{1.43e-07} & $2.8\times$ \\
\midrule
\multirow{3}{*}{Energy Dev.} & Unconst. & 5.17e-04 & 2.07e-04 & 1.04e-04 & \\
 & PINNs & 8.84e-05 & 3.58e-05 & 1.80e-05 & \\
 & Ours & \textbf{8.28e-05} & \textbf{3.36e-05} & \textbf{1.69e-05} & $1.1\times$ \\
\bottomrule
\end{tabular}
\end{table}

\subsection{Extended Pendulum (Q6: Scalability)}
\label{app:q6_results}

The extended pendulum tests whether compiled constraints remain effective as the underlying structure becomes more complex. The system requires preservation of both a Hamiltonian and a Casimir invariant in a state-dependent Poisson structure.

\begin{table}[H]
\centering
\caption{Extended pendulum (\textbf{Q6}: scalability with multiple simultaneous invariants). IF at $10\times$ relative to best baseline.}
\label{tab:results_pendulum}
\footnotesize
\setlength{\tabcolsep}{3pt}
\begin{tabular}{@{}llcccc@{}}
\toprule
\textbf{Metric} & \textbf{Model} & \textbf{$\times 2$} & \textbf{$\times 5$} & \textbf{$\times 10$} & \textbf{IF ($\times 10$)} \\
\midrule
\multirow{3}{*}{MSE (Train)} & Unconst. & 1.68e-07 & 1.68e-07 & 1.68e-07 & \\
 & PINNs & 1.32e-05 & 1.32e-05 & 1.32e-05 & \\
 & Ours & \textbf{9.71e-08} & \textbf{9.71e-08} & \textbf{9.71e-08} & $1.7\times$ \\
\midrule
\multirow{3}{*}{MSE (Extrap)} & Unconst. & 2.91e-06 & 3.30e-05 & 1.29e-04 & \\
 & PINNs & 1.13e-04 & 2.19e-03 & 5.68e-02 & \\
 & Ours & \textbf{5.15e-07} & \textbf{1.62e-06} & \textbf{3.23e-06} & $\mathbf{40\times}$ \\
\midrule
\multirow{3}{*}{MSE (Total)} & Unconst. & 1.54e-06 & 2.64e-05 & 1.16e-04 & \\
 & PINNs & 6.33e-05 & 1.75e-03 & 5.11e-02 & \\
 & Ours & \textbf{3.06e-07} & \textbf{1.32e-06} & \textbf{2.92e-06} & $\mathbf{40\times}$ \\
\midrule
\multirow{3}{*}{Casimir Dev.} & Unconst. & 9.68e-04 & 1.08e-03 & 1.07e-03 & \\
 & PINNs & 2.51e-03 & 7.73e-03 & 2.83e-02 & \\
 & Ours & \textbf{2.29e-04} & \textbf{4.84e-04} & \textbf{9.58e-04} & $1.1\times$ \\
\midrule
\multirow{3}{*}{Energy Dev.} & Unconst. & 7.48e-04 & 8.48e-04 & \textbf{8.37e-04} & \\
 & PINNs & 2.13e-03 & 6.61e-03 & 2.44e-02 & \\
 & Ours & \textbf{2.31e-04} & \textbf{4.96e-04} & 9.88e-04 & --- \\
\bottomrule
\end{tabular}
\end{table}

\section{Compiler Prompt Specification}

Table~\ref{tab:prompt-taxonomy} focuses on (1) how each invariant type is specified to the compiler (prompt specification) and
(2) the structure-preserving rewrite that is activated (enforced by construction).

\begin{table*}[htbp]
\centering
\caption{Prompt Taxonomy for the Invariant Compiler. }
\label{tab:prompt-taxonomy}
\small
\setlength{\tabcolsep}{6pt}
\renewcommand{\arraystretch}{1.2}
\begin{tabularx}{\textwidth}{@{}
p{0.14\textwidth}   
p{0.38\textwidth}    
Y                   
@{}}
\toprule
\textbf{Invariant Type} &
\textbf{Prompt Specification} &
\textbf{Compiler Rewrite} \\
\midrule

Simplex / Norm &
State lives on a simplex (mass conservation + non-negativity). Require exact preservation without penalties, normalization, or clamping. &
Compile to a norm-preserving latent representation (e.g., $\sqrt{\cdot}$ embedding) and enforce tangency via skew-symmetric generators; recover physical states by squaring. \\

Lorentz Cone &
State must remain in a Lorentz/convex cone for all $t$ (forward invariance). &
Enforce viability by projecting the vector field onto the tangent cone at the current state (closed-form cases for interior/boundary/apex). \\

PSD Cone &
Matrix-valued state must remain positive semidefinite ($\succeq 0$) throughout evolution. &
Reparameterize using a Cholesky factorization $P=LL^\top$ and evolve $L$; PSD is guaranteed by construction of $LL^\top$. \\

Center of Mass &
Center-of-mass position and total momentum must remain zero (mass-weighted constraints). &
Project out the center-of-mass components by mean subtraction (mass-weighted) in both velocity and acceleration updates. \\

Stoichiometric &
Linear conservation laws must hold (e.g., elemental/mass balances encoded by a matrix). &
Restrict dynamics to the null space of the conservation matrix via $\dot c = B\,r_\theta(c,t)$ with $\mathbf{M}B=0$, guaranteeing conservation for any $r_\theta$. \\

Hamiltonian &
Preserve energy and Poisson structure (Jacobi identity). &
Compile into canonical latent coordinates and impose structured Hamiltonian/Poisson dynamics (canonical $J$); induce the physical Poisson tensor via coordinate transformation. \\

Port-Hamiltonian &
Energy is non-increasing ($dH/dt\le 0$) while retaining Hamiltonian structure. &
Decompose into conservative and dissipative parts $\dot z=(J-R)\nabla K$ with $J^\top=-J$ and $R=LL^\top\succeq 0$, ensuring passive stability. \\

GENERIC &
Energy conservation and monotone entropy production with degeneracy conditions. &
Impose GENERIC decomposition with (i) Casimir-dependent entropy and (ii) projected dissipation enforcing degeneracy constraints; guarantees $dH/dt=0$ and $dS/dt\ge 0$. \\

First Integral &
Unknown/learned conserved quantities must be preserved exactly along trajectories. &
Learn invariants $V_i$ (upstream or jointly) and project the base field onto the tangent space of the learned constraint manifold using a (pseudo-)inverse Jacobian projector. \\

\bottomrule
\end{tabularx}
\end{table*}

\begin{lstlisting}[
  caption={Invariant Compiler Prompt (Specification)},
  label={lst:prompt},
  frame=none,
  backgroundcolor=\color{gray!8}
]
SYSTEM:
You are an invariant compiler for Neural ODEs.

Your task is to generate executable PyTorch code for a Neural ODE whose
trajectories satisfy the specified invariants BY CONSTRUCTION (up to
numerical integration error). Do not use penalty terms or post-hoc
projection unless explicitly requested.

USER:
GOAL
Generate PyTorch code for invariant-preserving Neural ODE.

INPUT SPECIFICATION
1) Base ODE (reference specification)
- State: x(t) in R^n with named components {x1, ..., xn}
- Dynamics: dx/dt = f(x,t (Symbolic form or function signature; may violate invariants)

2) Invariants (specified or discovered)
For each invariant:
- Invariant type (e.g., simplex, conservation law, Hamiltonian, PSD, first integral)
- Constraint equation(s) or inequality(ies)
  (e.g., sum_i x_i = 1, x_i >= 0)
- Any constants, domains, or initial-condition requirements

COMPILER REQUIREMENTS
- Select an appropriate geometric intermediate representation for each invariant type
- Rewrite the vector field into a structure-preserving form tangent to the admissible state manifold
- Parameterize only unconstrained submodules with neural networks; invariant structure must remain exact

OUTPUT REQUIREMENTS
Return a single, self-contained PyTorch implementation that includes:
- A vector field module (class VectorField(nn.Module))
- forward(t, x) returning dx/dt
- An example ODE solver call (e.g., torchdiffeq.odeint)
- Invariant diagnostics (e.g., violation along trajectories)
- A minimal demo showing invariant error at numerical precision

IMPORTANT
- Do not enforce invariants via penalties, normalization, or clamping
- Do not use post-hoc projection unless explicitly requested
- Return ONLY executable Python code
\end{lstlisting}

\end{document}